\documentclass{article}
\usepackage{microtype}
\usepackage{graphicx}
\usepackage{subfigure}
\usepackage{booktabs} 

\usepackage{hyperref}

\usepackage[accepted]{icml2024}

\usepackage{amsmath}
\usepackage{amssymb}
\usepackage{mathtools}
\usepackage{amsthm}

\usepackage[capitalize,noabbrev]{cleveref}

\theoremstyle{plain}
\newtheorem{theorem}{Theorem}[section]
\newtheorem{proposition}[theorem]{Proposition}

\theoremstyle{definition}

\theoremstyle{remark}
\newtheorem{remark}[theorem]{Remark}

\usepackage[textsize=tiny]{todonotes}

\usepackage{color}
\usepackage{hyperref}
\usepackage{bm}
\usepackage{nicefrac}
\usepackage{amssymb}
\usepackage{multirow}
\usepackage{multicol}
\usepackage{array}
\usepackage{amsthm}
\DeclareMathOperator*{\argmax}{arg\,max} 
\DeclareMathOperator*{\argmin}{arg\,min} 
\usepackage{subfigure}
\usepackage{makecell}
\usepackage{amssymb}

\newcommand{\E}{\mathbb{E}}
\newcommand{\eat}[1]{}

\icmltitlerunning{Efficient Black-box Adversarial Attacks via Bayesian Optimization Guided by a Function Prior}

\begin{document}

\twocolumn[
\icmltitle{Efficient Black-box Adversarial Attacks via Bayesian Optimization \\Guided by a Function Prior}



\icmlsetsymbol{equal}{*}

\begin{icmlauthorlist}
\icmlauthor{Shuyu Cheng}{equal,yyy,jq}
\icmlauthor{Yibo Miao}{equal,amss,ucas}
\icmlauthor{Yinpeng Dong}{yyy,rrr}
\icmlauthor{Xiao Yang}{yyy}
\icmlauthor{Xiao-Shan Gao}{amss,ucas,kaiyuan}
\icmlauthor{Jun Zhu}{yyy,rrr}
\end{icmlauthorlist}

\icmlaffiliation{yyy}{Dept. of Comp. Sci. and Tech., Institute for AI, Tsinghua-Bosch Joint ML Center, THBI Lab,
 BNRist Center, Tsinghua University, Beijing, 100084, China}
\icmlaffiliation{amss}{Academy of Mathematics and Systems Science, Chinese Academy of Sciences, Beijing, 100190, China}
\icmlaffiliation{ucas}{University of Chinese Academy of Sciences, Beijing, 100049, China}
\icmlaffiliation{kaiyuan}{Kaiyuan International Mathematical Sciences Institute}
\icmlaffiliation{rrr}{RealAI}
\icmlaffiliation{jq}{JQ Investments}

\icmlcorrespondingauthor{Yinpeng Dong}{dongyinpeng@tsinghua.edu.cn}
\icmlkeywords{Machine Learning, ICML}

\vskip 0.3in
]



\printAffiliationsAndNotice{\icmlEqualContribution} 

\begin{abstract}
This paper studies the challenging black-box adversarial attack that aims to generate adversarial examples against a black-box model by only using output feedback of the model to input queries. Some previous methods improve the query efficiency by incorporating the gradient of a surrogate white-box model into query-based attacks due to the adversarial transferability. However, the localized gradient is not informative enough, making these methods still query-intensive. In this paper, we propose a Prior-guided Bayesian Optimization (P-BO) algorithm that leverages the surrogate model as a \emph{global function prior} in black-box adversarial attacks. As the surrogate model contains rich prior information of the black-box one, P-BO models the attack objective with a Gaussian process whose mean function is initialized as the surrogate model's loss. Our theoretical analysis on the regret bound indicates that the performance of P-BO may be affected by a bad prior. Therefore, we further propose an adaptive integration strategy to automatically adjust a coefficient on the function prior by minimizing the regret bound. Extensive experiments on image classifiers and large vision-language models demonstrate the superiority of the proposed algorithm in reducing queries and improving attack success rates compared with the state-of-the-art black-box attacks.
Code is available at \url{https://github.com/yibo-miao/PBO-Attack}.
\end{abstract}

\section{Introduction}
A longstanding problem of deep learning is the vulnerability to adversarial examples \cite{Szegedy2013,Goodfellow2014}, which are generated by imposing small perturbations to natural examples but can mislead the target model. To identify the weaknesses of deep learning models and evaluate their robustness, adversarial attacks are widely studied to generate the worst-case adversarial examples. Some methods \cite{Goodfellow2014,Kurakin2016,Carlini2016} perform gradient-based optimization to maximize the classification loss, which inevitably requires access to the model architecture and parameters, known as \emph{white-box attacks}. On the other hand, \emph{black-box attacks} \cite{papernot2016practical} assume limited knowledge of the target model, which are more practical in real-world applications.

\begin{figure*}[t]
\centering
\includegraphics[width=0.92\textwidth]
{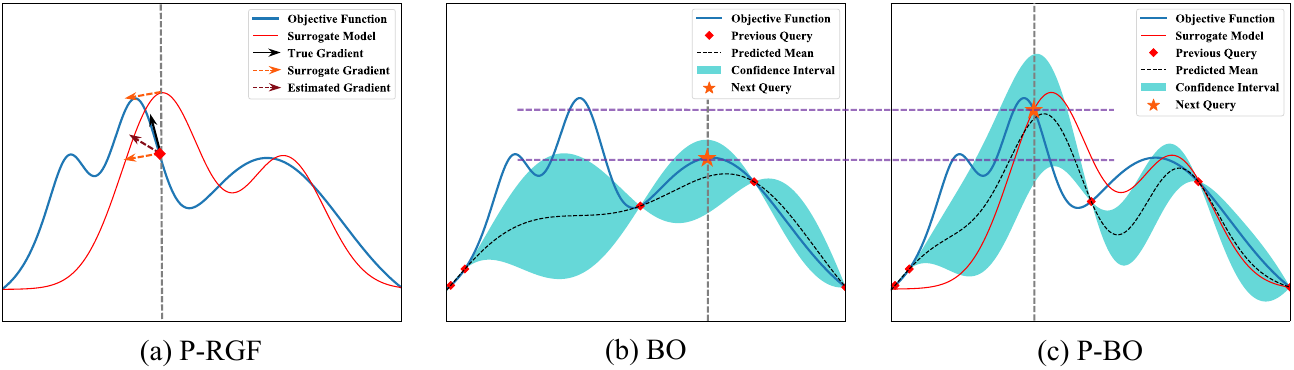}
\vspace{-1ex}
\caption{An illustration of the Prior-guided Random Gradient-Free (P-RGF) \cite{dong2022query}, Bayesian Optimization (BO), and Prior-guided Bayesian Optimization (P-BO) algorithms. The previous approaches, exemplified by P-RGF, adopt a local gradient of the surrogate model for gradient estimation. BO typically employs a zero-mean Gaussian process to approximate the unknown objective function, without leveraging any prior information. Our proposed P-BO algorithm integrates the surrogate model as a function prior into BO, which can better approximate the objective function and thus improve the query efficiency of black-box adversarial attacks.}
\label{fig:fig1}
\vspace{-1ex}
\end{figure*}

Tremendous efforts have been made to develop black-box adversarial attacks~\cite{papernot2016practical,Chen2017ZOO,Brendel2018Decision,Dong2017,ilyas2018black,Bhagoji_2018_ECCV,tu2018autozoom,ilyas2018prior,dong2022query},  
which can be generally categorized into \emph{transfer-based} attacks and \emph{query-based} attacks.
In transfer-based attacks, adversarial examples generated for a surrogate model are probable to fool the target model based on the adversarial transferability~\cite{Papernot20162,Liu2016}. 
Despite the recent improvements \cite{Dong2017,dong2019evading,xie2019improving,NI}, the success rate of transfer-based attacks is still limited for diverse models. This is attributed to the inherent dependence on the unknown similarity between the surrogate model and the target model, lacking an adjustment process. 
Differently, query-based attacks generate adversarial examples by leveraging the query feedback of the black-box model. The most prevalent approaches involve estimating the true gradient through zeroth-order optimization \cite{Chen2017ZOO,Bhagoji_2018_ECCV,tu2018autozoom,ilyas2018black}. There are also heuristic algorithms that do not rely on gradient estimation \cite{alzantot2019genattack,guo2019simple,al2019sign,andriushchenko2020square}. The main limitation of these methods is that they inevitably require a tremendous number of queries to perform a successful attack.

To improve the attack success rate and query efficiency, several methods~\cite{cheng2019improving,guo2019subspace,du2019query,yang2020learning,dong2022query} have been proposed to integrate transfer-based attacks with query-based attacks to have the best of both worlds. They typically leverage the input gradient of a surrogate white-box model as the transfer-based prior to improve query-based optimization.
However, the surrogate gradient is localized and may not be informative enough, as illustrated in Fig.~\ref{fig:fig1}(a).
These previous methods are unable to sufficiently exploit the global information of the surrogate model as a \emph{function prior} in the entire space. 
Consequently, they still require hundreds of queries to successfully attack the target model.
Intuitively, the function prior can provide more abundant information of the black-box model, thus can further improve the black-box attack performance if appropriately utilized. 
Therefore, we explore how to leverage the \emph{global function prior} instead of the \emph{local gradient prior} for improving the efficiency of black-box attacks in this paper.

Bayesian Optimization (BO) \cite{jones1998efficient} is a classic black-box optimization approach of finding global optima of the objective functions, which can seamlessly integrate prior information over functions. 
In recent years, several studies \cite{zhao2019design, shukla2019black, ru2019bayesopt} have applied Bayesian optimization to black-box attacks.
These methods adopt a Bayesian statistical
model (e.g., Gaussian process) 
to approximate the attack objective and update a posterior distribution according to which the next point to query is chosen, as shown in Fig.~\ref{fig:fig1}(b).
Although these methods are effective with low query budgets, they usually adopt a zero-mean Gaussian process, which does not leverage any prior information, leaving room for improvement.

To address the aforementioned issues and improve black-box attacks, we propose a \textbf{Prior-guided Bayesian Optimization (P-BO)} algorithm, which integrates the surrogate model \footnote{To avoid ambiguity, in this paper we use the term ``surrogate model'' to indicate the white-box model in attacks, rather than the Bayesian statistical model (also called statistical surrogate) in BO.} as a function prior into Bayesian optimization, as shown in Fig.~\ref{fig:fig1}(c). 
Specifically, P-BO initializes the mean function of the Gaussian process with the surrogate model's loss and updates the posterior distribution given the observed values of the objective function at the sampled locations.
We theoretically analyze the regret bound of P-BO, which is closely related to the optimization error and convergence speed of the algorithm. 
Based on the theoretical analysis, we notice that the straightforward integration of the surrogate model into the Gaussian process may lead to a worse regret bound.
Therefore, we further propose an \emph{adaptive integration} strategy, which sets an adjustable coefficient on the surrogate model controlling the strength of utilizing the function prior. The optimal value of the coefficient is determined according to the quality of the prior by minimizing the regret bound. 
With this technique, P-BO can largely prevent performance degradation when the function prior is useless. 

We conduct extensive experiments on CIFAR-10 and ImageNet to confirm the superiority of our method. The results demonstrate that P-BO significantly outperforms the previous state-of-the-art black-box attacks in terms of the  success rate and query efficiency. For example, P-BO needs less than 20 queries on average to obtain 100\% attack success rates on CIFAR-10, greatly outperforming the existing methods. Furthermore, we conduct experiments on large vision-language models (VLMs)  to validate the effectiveness and practicality of our method for attacking the prevailing multimodal foundation models.

\vspace{-0.5ex}
\section{Preliminaries}

In this section, we introduce the preliminary knowledge of black-box attacks and Bayesian optimization. More discussions on related work can be found in Appendix \ref{app:b}.

\subsection{Black-box Adversarial Attacks}

Given a natural input $\bm{x}^{nat}\in\mathbb{R}^D$ and its ground-truth class $c$, adversarial attack aims to generate an adversarial example $\bm{x}^{adv}$ by solving a constrained optimization problem: 
\begin{equation}
    \bm{x}^{adv} = \argmax_{\bm{x}\in A} f(\bm{x},c),
    \label{eq:problem}
\end{equation}
where $f$ is an attack objective function on top of the target model (e.g., cross-entropy loss, CW loss \cite{Carlini2016}), $A = \{\bm{x}|\|\bm{x}-\bm{x}^{nat}\|_\infty \leq \epsilon\}$ is the allowed space of adversarial examples and we consider the $\ell_\infty$ norm with the perturbation budget $\epsilon$ in this paper. In the following, we omit the class $c$ in $f$ for simplicity.

Several white-box attacks \cite{Goodfellow2014,Kurakin2016,Carlini2016,Madry2017Towards} have been proposed to solve problem \eqref{eq:problem} by gradient-based optimization.
The typical projected gradient descent method (PGD)~\cite{Madry2017Towards} performs iterative update as 
\begin{equation}
\bm{x}_{t+1}^{adv} = \Pi_A (\bm{x}_t^{adv} + \eta\cdot \mathrm{sign}(\nabla_{\bm{x}}f(\bm{x}_t^{adv}))),
\label{eq:iter}
\end{equation}
where $\Pi_A$ projects the adversarial example onto the $\ell_\infty$ ball around $\bm{x}^{nat}$ with radius $\epsilon$, $\eta$ is the step size, and $\mathrm{sign}(\cdot)$ is the sign function to normalize the gradient.

On the other hand, black-box attacks assume limited knowledge about the target model, which can be challenging yet practical in various real-world applications.
Black-box attacks can be roughly divided into transfer-based attacks and query-based attacks. 
Transfer-based attacks craft adversarial examples for a surrogate model $f'$, which are probable to fool the black-box model $f$ due to the transferability \cite{Papernot20162,Liu2016}. 
Some query-based attacks estimate the gradient $\nabla_{\bm{x}}f(\bm{x})$ by zeroth-order optimization methods, when the loss values could be obtained through model queries. 
For example, the random gradient-free (RGF) method \cite{ghadimi2013stochastic,duchi2015optimal,nesterov2017random} estimates the gradient as
\begin{equation}
     \nabla_{\bm{x}}f(\bm{x}) \approx \frac{1}{q}\sum_{i=1}^{q}\frac{f(\bm{x}+\sigma \bm{u}_i)-f(\bm{x})}{\sigma}\cdot \bm{u}_i,
\label{eq:estimate}
\end{equation}
where $\{\bm{u}_i\}_{i=1}^q$ are random directions.
Query-based attacks usually demand hundreds of or thousands of queries to successfully generate an adversarial example due to the high-dimensional search space.

There is also plenty of work \cite{cheng2019improving,guo2019subspace,du2019query,yang2020learning,ma2020switching,huang2020black,dong2022query,yin2023generalizable} integrating transfer-based attacks with query-based attacks to achieve high attack success rate and high query efficiency simultaneously. One straightforward idea is to utilize the gradient of the surrogate model as a transfer-based prior to obtain a more accurate gradient estimate
\cite{cheng2019improving,dong2022query} or restrict the search space
spanned by the surrogate gradient(s) \cite{guo2019subspace,ma2020switching,yang2020learning}.
Although these methods are effective in expediting convergence and reducing the number of queries, the surrogate gradient is localized and can be misleading, limiting their effectiveness. Besides, other methods learn a generalizable model-based prior based on the surrogate model \cite{du2019query,huang2020black,yin2023generalizable}. But these methods require an additional dataset to train the attack generator, which is not applicable when the data is scarce.
In this paper, we aim to develop an efficient and elegant black-box attack to leverage the global function prior of the surrogate model without additional data.

\subsection{Bayesian Optimization}

Bayesian Optimization (BO)~\cite{jones1998efficient} is an efficient method for finding global optima of black-box optimization problems. BO consists of two key components: a Bayesian statistical model, such as a Gaussian process (GP) \cite{rasmussen2003gaussian}, which approximates the unknown objective function $f$; and an acquisition function $\alpha(\cdot)$, which is maximized to recommend the next query location by balancing exploitation and exploration with the most promising expected improvement in function values.

Specifically, 
assume $f$ a priori follows the Gaussian process with mean $0$ (which is common in previous work \cite{frazier2018tutorial}) and kernel function $k(\cdot,\cdot)$, denoted as $f\sim \operatorname{GP}(0, k)$. Given the observation data $\mathcal{D}=\{(\bm{x}_1, y_1),\ldots, (\bm{x}_T, y_T)\}$ where $y_i = f(\bm{x}_i)$, the predictive posterior distribution for $f(\bm{x})$ at a test point $\bm{x}$, denoted as $p(f(\bm{x})|\bm{x};\mathcal{D})$, follows a Gaussian distribution $\mathcal{N}(\mu_T(\bm{x}), \sigma_T^2(\bm{x}))$, where
\begin{align}
    \mu_T(\bm{x})&=\mathbf{k}_T(\bm{x})^\top \mathbf{K}_T^{-1} \mathbf{y}_T, \label{eq:gp-mean}\\
    \sigma_T^2(\bm{x})&=k(\bm{x}, \bm{x}) - \mathbf{k}_T(\bm{x})^\top \mathbf{K}_T^{-1} \mathbf{k}_T(\bm{x}), \label{eq:gp-variance}
\end{align}
where $\mathbf{k}_T(\bm{x})=[k(\bm{x}_1,\bm{x}),k(\bm{x}_2,\bm{x}),\ldots,k(\bm{x}_T,\bm{x})]^\top$, $\mathbf{K}_T$ 
is a $T\times T$ matrix with its $(i, j)$-th element being $k(\bm{x}_i, \bm{x}_j)$, and $\mathbf{y}_T=[y_1,\ldots,y_T]^\top$. 

Based on the posterior distribution, we can construct the acquisition function $\alpha({\cdot})$, 
such as Expected Improvement (EI) \cite{movckus1975bayesian}, Probability of Improvement (PI) \cite{jones2001taxonomy}, Upper-Confidence Bounds (UCB) \cite{srinivas2009gaussian}, and entropy-based methods \cite{hernandez2014predictive}.
In this work, we choose UCB as the acquisition function to analyze the regret bound, which is expressed as
\begin{align}
\alpha(\bm{x})=\mu_T(\bm{x})+\beta \sigma_T(\bm{x}), \label{eq:ucb}
\end{align}
where 
the coefficient $\beta$ balances exploration (i.e., encouraging queries in regions with high predictive variance) and exploitation (i.e., encouraging queries in regions with high predictive mean).
Subsequently, the next point $\bm{x}_{T+1}$ is selected by maximizing the acquisition function as $\bm{x}_{T+1}=\argmax_{\bm{x}\in A} \alpha(\bm{x})$ and then used to query the objective $f$.

In recent years, several studies have applied BO to black-box attacks. The pioneering work \cite{suya2017query} initially applies BO to attacking spam email classifiers. Subsequently, \citet{zhao2019design, shukla2019black, ru2019bayesopt, miao2022isometric} extend the application of BO to attacking deep models, providing empirical evidence of its effectiveness. However, to date, there is a notable absence of research integrating prior information into BO for black-box attacks.
Beyond black-box attacks, some methods 
\cite{souza2021bayesian,hvarfner2022pi} try different ways of integrating prior information into BO, but they do not utilize a deterministic function prior as in our work,
and they are not applied to black-box attacks (see Appendix \ref{app:b} for details). 

\vspace{-1ex}
\section{Methodology}
\vspace{-0.5ex}

In this section, we introduce the Prior-guided Bayesian Optimization (P-BO) algorithm and the adaptive integration strategy in Sec.~\ref{sec:3-1} and Sec.~\ref{sec:3-2}, respectively.

\vspace{-0.5ex}
\subsection{Prior-guided Bayesian Optimization}\label{sec:3-1}

As discussed above, our main motivation is to improve the efficiency of black-box attacks by leveraging the global function prior of a surrogate model. As Bayesian optimization (BO) enables global optimization of the black-box objective function by building a probabilistic model, it can seamlessly integrate prior information over functions. Therefore, we propose a novel  \textbf{Prior-guided Bayesian Optimization (P-BO)} algorithm for more efficient black-box attacks.

Specifically, we consider optimizing the objective function $f(\bm{x})$ in Eq.~\eqref{eq:problem} for attacking a black-box model. We assume that we have a surrogate white-box model and can obtain its objective function $f'$ in the same form of $f$. 
Due to the similarity between the surrogate model and the black-box one, $f'$ could exhibit some similarities to $f$ in the function space. And that is why adversarial examples generated by optimizing $f'$ can have a certain probability to also mislead $f$ (i.e., adversarial transferability) \cite{papernot2016practical,Liu2016,Dong2017}. Therefore, we regard $f'$ as a function prior to optimize the black-box objective $f$. P-BO initializes the mean function of the Gaussian process with the function prior $f'$, so $f$ a priori follows $f \sim \operatorname{GP}(f',k)$. Similar to Eq.~\eqref{eq:gp-mean} and Eq.~\eqref{eq:gp-variance}, the posterior distribution $p(f(\bm{x})|\bm{x};\mathcal{D})$ given the observation data $\mathcal{D}$ also follows a Gaussian distribution 
$\mathcal{N}(\mu_T(\bm{x}), \sigma_T^2(\bm{x}))$, where
\begin{align}
    \mu_T(\bm{x})&=\mathbf{k}_T(\bm{x})^\top \mathbf{K}_T^{-1} (\mathbf{y}_T-\mathbf{y}'_T) + f'(\bm{x}), \label{eq:gp-mean-mod}\\
    \sigma_T^2(\bm{x})&=k(\bm{x}, \bm{x}) - \mathbf{k}_T(\bm{x})^\top \mathbf{K}_T^{-1} \mathbf{k}_T(\bm{x}). \label{eq:gp-variance-mod}
\end{align}
where $\mathbf{k}_T(\bm{x})$, $\mathbf{K}_T$, $\mathbf{y}_T$ are the same with those in Eq.~\eqref{eq:gp-mean} and Eq.~\eqref{eq:gp-variance}, and $\mathbf{y}'_T=[f'(\bm{x}_1),\ldots,f'(\bm{x}_T)]^\top$.

By comparing Eq.~\eqref{eq:gp-mean-mod}-\eqref{eq:gp-variance-mod} of P-BO with Eq.~\eqref{eq:gp-mean}-\eqref{eq:gp-variance} of BO, we notice that only the mean $\mu_T(\bm{x})$ changes after the function prior is introduced. 
Hence P-BO can be viewed as first modeling the residual $f-f'$ using its observed values with a zero-mean Gaussian process $\operatorname{GP}(0,k)$, and then adding $f'$ to form the model of $f$. Intuitively, the approximation error of $f$ becomes dependent on the magnitude of $f-f'$. When $f=f'$, we could extract an accurate estimation of $f$ without the requirement of observed values since $\mu_T=f$.
It is reasonable to anticipate that when $f'$ is closer to $f$, the performance of P-BO will be better.

In the following, we theoretically analyze the regret bound of our proposed P-BO algorithm. 
Following \citet{srinivas2009gaussian}, we define the instantaneous regret $r_t=f(\bm{x}^*)-f(\bm{x}_t)$ at the $t$-th iteration, where $\bm{x}^*=\argmax_{\bm{x}\in A}f(\bm{x})$ is the global maximum. The cumulative regret $R_T$ after $T$ iterations is $R_T=\sum_{t=1}^T r_t$. In the realm of black-box optimization, we are typically concerned with the optimization error $\min_{1\leq t\leq T} r_t$, which can be upper bounded by $\frac{R_T}{T}$. Therefore, the convergence rate of the optimization algorithm is closely related to the regret $R_T$. Below, we derive the regret bound of P-BO with the statistical model $\operatorname{GP}(f',k)$ and the UCB acquisition function.
For technical reasons, we shall consider an observation noise $\mathcal{N}(0, \sigma^2)$ in the Gaussian process modeling procedure (replacing $\mathbf{K}_T$ in Eq.~\eqref{eq:gp-mean-mod}-\eqref{eq:gp-variance-mod} with $\mathbf{K}_T+\sigma^2\mathbf{I}$, see e.g. Theorem 3.1 of \citet{kanagawa2018gaussian}), but the actual observed value of $f$ is deterministic and without noise.

\begin{theorem}\label{thm:1}
(Proof in Appendix \ref{app:a-1}) Assume $f$ and $f'$ lie in the Reproducing Kernel Hilbert Space (RKHS) corresponding to kernel $k$, and let $\|\cdot\|_k$ denote the RKHS norm. In Bayesian optimization, suppose we model $f$ by $\operatorname{GP}(f',k)$ with observation noise $\mathcal{N}(0, \sigma^2)$, and we use the UCB acquisition function defined in Eq.~\eqref{eq:ucb} where $\beta=\|f-f'\|_k$. 
Then the regret $R_T$ satisfies
\begin{align}
    R_T\leq \|f-f'\|_k\sqrt{\frac{8}{\log (1+\sigma^{-2})}T\gamma_T},
\end{align}
where $\gamma_T=\frac{1}{2}\max_{\bm{x}_1,\ldots,\bm{x}_T\in A}\log |\mathbf{I}+\sigma^{-2}\mathbf{K}_T|$ in which $\mathbf{K}_T$ 
is the covariance matrix with its $(i, j)$-th element being $k(\bm{x}_i, \bm{x}_j)$.
\end{theorem}

\begin{remark}\label{remark:1}
Intuitively, incorporating the function prior $ f' $ into the modeling of $f$ using the Gaussian process $\operatorname{GP}(f', k)$ involves substituting $\|f\|_k$ with $\|f-f'\|_k$ in the upper bound of the regret $R_T$ in BO \cite{srinivas2009gaussian}. Consequently, when $f'\approx f$, employing the Gaussian process $\operatorname{GP}(f', k)$ to model $ f $ significantly lowers the regret.
\end{remark}

Theorem~\ref{thm:1} indicates that the regret bound is proportional to $\|f-f'\|_k$. 
To achieve better performance of using P-BO, we desire a small value of $\|f-f'\|_k$, at least satisfying $\|f-f'\|_k \leq \|f\|_k$. Although this requirement holds when $f'$ approximates $f$ well, for functions defined in a high-dimensional space, this requirement is quite strong. In Appendix \ref{app:c}, we provide an example showing that when $f$ and $f'$ are both random linear functions, $\|f-f'\|_k > \|f\|_k$ with high probability. 
This implies that when the target and prior functions are not closely aligned, directly modeling $f$ with $\operatorname{GP}(f', k)$ could usually degrade the performance of P-BO compared with the vanilla BO. Therefore, we further propose an adaptive integration strategy, as detailed next.

\subsection{Adaptive Integration Strategy}\label{sec:3-2}

To address the aforementioned problem, our fundamental idea is to set a coefficient $\lambda$ on the function prior and automatically adjust $\lambda$ according to the quality of the prior. We aim to ensure that the performance of P-BO would not be affected by a useless prior, and to optimize $\lambda$ to achieve better performance in the presence of a useful prior.

Based on our theoretical analysis in Theorem~\ref{thm:1}, we employ a straightforward approach by replacing $f'$ with $\lambda f'$, where $\lambda \in\mathbb{R}$ can be interpreted as the weight of integrating the function prior. Therefore, P-BO models the objective as $f\sim\operatorname{GP}(\lambda f', k)$, whose posterior distribution can be similarly derived. Note that $\lambda=0$ corresponds to the zero-mean Gaussian process and $
\lambda=1$ corresponds to the case presented in Sec.~\ref{sec:3-1}. The motivation behind this design is:
\begin{align}
    \argmin_\lambda \|f-\lambda f'\|_k^2=\|f\|_k^2-\frac{\langle f,f'\rangle_k^2}{\|f'\|_k^2}\leq \|f\|_k^2,
\end{align}
where the minimum is achieved at $\lambda^*=\frac{\langle f,f'\rangle_k}{\|f'\|_k^2}$, and $\langle\cdot,\cdot\rangle_k$ denotes the inner product in the RKHS corresponding to the kernel $k$. Thus, by selecting an appropriate coefficient $\lambda$, we can ensure that the regret bound in Theorem~\ref{thm:1} is reduced.

Although the optimal solution $\lambda^*=\frac{\langle f,f'\rangle_k}{\|f'\|_k^2}$ has an analytical form, the function $f$ is unknown in black-box optimization. The inner product and norm of functions in the RKHS are also difficult to compute accurately. Consequently, calculating the optimal coefficient $\lambda^*$ poses a significant challenge. 
To address this, we develop a heuristic approximation method based on the following insightful proposition to estimate the norm $\|f-\lambda f'\|_k$ in the RKHS. This estimate is then utilized to further determine the optimal coefficient $\lambda$ minimizing $\|f-\lambda f'\|_k^2$. 

\begin{proposition}\label{thm:2}
(Proof in Appendix \ref{app:a-2}) Given the function $\mu_T$ defined in Eq.~\eqref{eq:gp-mean}, which is the predictive posterior mean of the objective function modeled by $\operatorname{GP}(0,k)$ given observation data $\mathcal{D}$, and let $\mathcal{H}$ be the RKHS corresponding to the kernel $k$, then we have 
\begin{align}
\label{eq:interpolation}
    \mu_T = \argmin_{h\in\mathcal{H}} \|h\|_k,\ \ \mathrm{s.t.}\ \ \forall 1\leq t\leq T, h(\bm{x}_t)=y_t,
\end{align}
and $\|\mu_T\|_k^2=\mathbf{y}_T^\top \mathbf{K}_T^{-1} \mathbf{y}_T$.
\end{proposition}

\begin{remark}\label{remark:2}
$\mu_T$ could be viewed as the ``smoothest'' (in the sense of minimizing the RKHS norm) interpolation for the given dataset $\mathcal{D}$. Using its norm to estimate $\|h\|_k$, it can be regarded that $\|h\|_k^2\approx \|\mu_T\|_k^2 =  \mathbf{y}_T^\top \mathbf{K}_T^{-1} \mathbf{y}_T$. This estimate might be coarse when the size of $\mathcal{D}$ is small since it is a lower bound, but it is almost the best we can do, since giving an upper bound of $\|h\|_k$ with the only restriction $\mathcal{D}$ is intuitively infeasible without more assumptions.
\end{remark}

To optimize $\|f-\lambda f'\|_k^2$, 
whose values at $\bm{x}_1,\ldots,\bm{x}_T$ can be stacked into the vector $\mathbf{y}_T-\lambda \mathbf{y}_T'$, we can approximate $\|f-\lambda f'\|_k^2$ using Remark~\ref{remark:2} with $h=f-\lambda f'$:
\begin{align}
    \|f-\lambda f'\|_k^2&\approx (\mathbf{y}_T-\lambda \mathbf{y}_T')^\top \mathbf{K}_T^{-1} (\mathbf{y}_T-\lambda \mathbf{y}_T') \nonumber\\*
    &= -\log \mathcal{N}(\mathbf{y}_T|\lambda \mathbf{y}_T', \mathbf{K}_T) + \mathrm{const} \\*
    &= -\log p(\mathcal{D}|\lambda) + \mathrm{const}, \nonumber
\end{align}
where the $\mathrm{const}$ term is independent of $\lambda$, and $\log p(\mathcal{D}|\lambda)$ represents the log-likelihood of $\mathcal{D}$ under the modeling of $\operatorname{GP}(\lambda f',k)$. Notably, solving $\argmin_\lambda \|f-\lambda f'\|_k^2$ can be approximated as maximizing the log-likelihood, i.e., solving $\argmax_\lambda\log p(\mathcal{D}|\lambda)$. Therefore, for the sake of convenience, in implementation we can treat the integration coefficient $\lambda$ as a hyperparameter in the Gaussian process model, and adaptively adjust $\lambda$ together with other hyperparameters (e.g., the hyperparameters of the kernel) by maximizing the log-likelihood of the dataset.

\begin{algorithm}[t]
\small
\caption{Prior-guided Bayesian Optimization (P-BO)}
\label{alg:pg-bayesopt}
\begin{algorithmic}[1]
\STATE {\bfseries Input:} Objective function $f$; function prior $f'$; search space $A$; number of queries $N$; kernel function $k$; balancing coefficient $\beta$.
\STATE {\bfseries Output:} Approximate solution $\bm{x}^*$ maximizing $f$.
\STATE 
Construct an initial dataset: $\mathcal{D}\leftarrow \{(\bm{x}_1, y_1),\ldots,(\bm{x}_S, y_S)\}$;
\FOR{$T = S$ \textbf{to} $N-1$}
\STATE 
Normalize $\mathbf{y}_T$, $\mathbf{y}_T'$, and $f'$ (see text for details); 
\STATE 
Solve $\lambda^*$ by maximizing $\log p(\mathcal{D}|\lambda)$;
\STATE 
Compute posterior $p(f(\bm{x})|\bm{x};\mathcal{D})=\mathcal{N}(\mu_T(\bm{x}), \sigma_T^2(\bm{x}))$ of $\operatorname{GP}(\lambda f',k)$, where
\vspace{-2ex}
\begin{align*}
    \mu_T(\bm{x})&= \mathbf{k}_T(\bm{x})^\top \mathbf{K}_T^{-1} (\mathbf{y}_T-\lambda^*\mathbf{y}'_T) + \lambda^* f'(\bm{x}),
\\
    \sigma_T^2(\bm{x})&= k(\bm{x}, \bm{x}) - \mathbf{k}_T(\bm{x})^\top \mathbf{K}_T^{-1} \mathbf{k}_T(\bm{x});
\end{align*}
\vspace{-4ex}
\STATE 
Compute acquisition function: 
$\alpha(\bm{x})=\mu_T(\bm{x})+\beta \sigma_T(\bm{x})$;
\STATE 
Obtain the next query point: $\bm{x}_{T+1}= \operatorname{argmax}_{\bm{x}\in A} \alpha(\bm{x})$;
\STATE 
Query $f$ at $\bm{x}_{T+1}$ to obtain $y_{T+1} = f(\bm{x}_{T=1})$;
\STATE Update the dataset: $\mathcal{D}\leftarrow \mathcal{D} \cup \{(\bm{x}_{T+1}, y_{T+1})\}$;
\ENDFOR
\STATE 
\textbf{return} $\bm{x}_N$.
\end{algorithmic}
\end{algorithm}

\begin{table*}[!t]
\vspace{-2ex}
  \caption{The experimental results of black-box attacks against DenseNet-121, ResNet-50, and SENet-18 under the $\ell_\infty$ norm on CIFAR-10. We report the attack success rate (ASR) and the average/median number of queries (AVG. Q/MED. Q) needed to generate an adversarial example over successful attacks. We mark the best results in \textbf{bold}.}
  \setlength{\tabcolsep}{4pt}
  \label{tab:cifar-l8}
  \centering\small
  \begin{tabular}{l|ccc|ccc|ccc}
    \hline
    \multirow{2}{*}{Method} & \multicolumn{3}{c|}{ResNet-50} & \multicolumn{3}{c|}{DenseNet-121} & \multicolumn{3}{c}{SENet-18}\\
    \cline{2-10}
    & ASR & AVG. Q & MED. Q & ASR & AVG. Q & MED. Q & ASR & AVG. Q & MED. Q \\
    \hline
    NES~\cite{ilyas2018black} & 79.9\% & 353 & 306 & 75.5\% & 358 & 306 & 80.2\% & 335 & 255 \\
    Bandits\textsubscript{T}~\cite{ilyas2018prior} & 92.0\% & 214 & 130 & 90.7\% & 240 & 158 & 93.4\% & 203 & 128 \\
    $\mathcal{N}$ATTACK~\cite{li2019nattack} & 94.3\% & 296 & 204 & 93.9\% & 309 & 255 & 94.9\% & 270 & 204 \\
    SignHunter~\cite{al2019sign} & 84.4\% & 267 & 180 & 85.5\% & 258 & 158 & 86.1\% & 251 & 156 \\
    Square~\cite{andriushchenko2020square} & 94.2\% & 206 & 100 & 94.3\% & 216 & 124 & 94.9\% & 197 & 97 \\
    NPAttack~\cite{bai2023query} & 98.5\% & 151 & 75 & 96.0\% & 86 & 50 & 95.5\% & 124 & 50 \\
    RGF~\cite{cheng2019improving} & 86.9\% & 232 & 149 & 83.1\% & 254 & 179 & 87.6\% & 223 & 149 \\
    P-RGF~\cite{dong2022query} & 98.4\% & 55 & 25 & 95.8\% & 76 & 27 & 98.3\% & 49 & 23 \\
    BO~\cite{ru2019bayesopt} & 99.6\% & 83 & 44 & 99.7\% & 93 & 51 & 99.7\% & 81 & 41 \\
    P-BO ($\lambda=1$, \textbf{ours}) & 99.9\% & 16 & \bf11 & 99.7\% & 25 & \bf11 & 99.9\% & \bf14 & \bf11 \\
    P-BO ($\lambda^*$, \textbf{ours}) & \bf100.0\% & \bf15 & \bf11 & \bf100.0\% & \bf19 & \bf11 & \bf100.0\% & \bf14 & \bf11 \\
    \hline
  \end{tabular}
  \vspace{-2ex}
\end{table*}

In summary, the algorithm of P-BO is outlined in Alg.~\ref{alg:pg-bayesopt}. First, we construct an initial dataset $\mathcal{D}=\{(\bm{x}_i, y_i)\}_{i=1}^{S}$ by randomly sampling $\bm{x}_i$ in the search space $A$ and obtaining $y_i=f(\bm{x}_i)$. At each iteration, we normalize $\mathbf{y}_T$, $\mathbf{y}_T'$, and $f'$ to a good and similar numerical range: $\mathbf{y}_{T}\leftarrow \frac{\mathbf{y}_T-\mu}{\sigma}$, $\mathbf{y}_{T}'\leftarrow \frac{\mathbf{y}_T'-\mu'}{\sigma'}$, $f'\leftarrow \frac{f'-\mu'}{\sigma'}$, where 
$\mu$ and $\sigma$ represent the mean and standard deviation of $\mathbf{y}_T$, and $\mu'$ and $\sigma'$ represent the mean and standard deviation of $\mathbf{y}'_T$. Subsequently, we obtain an optimal integration coefficient $\lambda^*$ through maximum likelihood. Following this, we model $f$ by Gaussian process $\operatorname{GP}(\lambda^* f',k)$ and compute the posterior distribution $p(f(\bm{x})|\bm{x};\mathcal{D})$.
Based on the posterior distribution, we compute the acquisition function $\alpha(\bm{x})=\mu_T(\bm{x})+\beta \sigma_T(\bm{x})$ and maximize the acquisition function $\alpha(\bm{x})$ to obtain the next query point $\bm{x}_{T+1}$. Finally, we query the objective function $f$ to obtain $y_{T+1}=f(\bm{x}_{T+1})$ and update the dataset $\mathcal{D}$. In the algorithm, the maximization problems of finding $\lambda^*$ and $\bm{x}_{T+1}$ are solved by gradient-based methods \cite{frazier2018tutorial}.

\section{Experiments}

\begin{table*}
\vspace{-2ex}
  \caption{The experimental results of black-box attacks against Inception-v3, MobileNet-v2, and ViT-B/16 under the $\ell_\infty$ norm on ImageNet. We report the attack success rate (ASR) and the average/median number of queries (AVG. Q/MED. Q) needed to generate an adversarial example over successful attacks. We mark the best results in \textbf{bold}. The subscript ``D'' denotes the methods with dimensionality reduction.}
  \label{tab:imagenet-l8}
\setlength{\tabcolsep}{4.5pt}
  \centering\small
  \begin{tabular}{l|ccc|ccc|ccc}
    \hline
    \multirow{2}{*}{Method} & \multicolumn{3}{c|}{Inception-v3} & \multicolumn{3}{c|}{MobileNet-v2} & \multicolumn{3}{c}{ViT-B/16}\\
    \cline{2-10}
    & ASR & AVG. Q & MED. Q & ASR & AVG. Q & MED. Q & ASR & AVG. Q & MED. Q \\
    \hline
    NES~\cite{ilyas2018black} & 53.6\% & 348 & 306 & 61.5\% & 320 & 255 & 41.4\% & 339 & 255 \\
    Bandits\textsubscript{T}~\cite{ilyas2018prior} & 45.8\% & 300 & 209 & 67.4\% & 262 & 141 & 42.3\% & 303 & 200 \\
    $\mathcal{N}$ATTACK~\cite{li2019nattack} & 78.7\% & 347 & 255 & 87.1\% & 335 & 255 & 63.4\% & 352 & 306 \\
    SignHunter~\cite{al2019sign} & 80.3\% & 223 & 128 & 81.7\% & 252 & 156 & 73.0\% & 212 & 96 \\
    Square~\cite{andriushchenko2020square} & 76.8\% & 194 & 100 & 97.8\% & 95 & \bf17 & 71.0\% & 188 & \bf74 \\
    NPAttack~\cite{bai2023query} & 75.0\% & 381 & 300 & 86.9\% & 428 & 400 & 63.6\% & 374 & 300 \\
    RGF~\cite{cheng2019improving} & 54.5\% & 260 & 167 & 65.2\% & 206 & 107 & 44.2\% & 269 & 185 \\
    P-RGF~\cite{dong2022query} & 72.9\% & 186 & 99 & 84.9\% & 159 & 71 & 52.7\% & 222 & 129 \\
    BO~\cite{ru2019bayesopt} & 89.2\% & 169 & 105 & 97.8\% & 133 & 85 & 77.5\% & 219 & 140 \\
    P-BO ($\lambda=1$, \textbf{ours}) & 60.8\% & 186 & 75 & 78.7\% & 136 & 26 & 43.9\% & 223 & 115 \\
    P-BO ($\lambda^*$, \textbf{ours}) & 91.4\% & 115 & 58 & 98.7\% & 95 & 54 & 83.6\% & 189 & 93 \\
    \hline
    Bandits\textsubscript{TD}~\cite{ilyas2018prior} & 60.3\% & 260 & 142 & 76.5\% & 233 & 110 & 54.2\% & 267 & 148 \\
    RGF\textsubscript{D}~\cite{cheng2019improving} & 73.6\% & 205 & 107 & 69.8\% & 181 & 89 & 55.8\% & 232 & 149 \\
    P-RGF\textsubscript{D}~\cite{dong2022query} & 80.1\% & 194 & 113 & 87.5\% & 162 & 75 & 61.3\% & 236 & 145 \\
    BO\textsubscript{D}~\cite{ru2019bayesopt} & 94.1\% & 104 & 58 & 98.2\% & 102 & 61 & 86.7\% & 170 & 116 \\
    P-BO\textsubscript{D} ($\lambda=1$, \textbf{ours}) & 85.4\% & 182 & 90 & 86.8\% & 193 & 56 & 67.5\% & 236 & 240 \\
    P-BO\textsubscript{D} ($\lambda^*$, \textbf{ours}) & \bf94.4\% & \bf81 & \bf45 & \bf98.8\% & \bf94 & 60 & \bf88.2\% & \bf148 & 81 \\
    \hline
  \end{tabular}
  \vspace{-2.5ex}
\end{table*}

In this section, we present the empirical results to demonstrate the effectiveness of the proposed methods on attacking black-box models. We perform untargeted attacks under the $\ell_{\infty}$ norm on CIFAR-10 \cite{krizhevsky2009learning} and ImageNet \cite{russakovsky2015imagenet} for image classifiers, and MS-COCO \cite{lin2014microsoft} for large Vision-Language Models (VLMs). We first specify the experimental setting in Sec.~\ref{sec:4-1}. Then we show the results on CIFAR-10 in Sec.~\ref{sec:4-2}, ImageNet in Sec.~\ref{sec:4-3}, and VLMs in Sec.~\ref{sec:4-4}, respectively.
We also conduct experiments on defense models in Sec.~\ref{sec:4-5} and show the performance of adaptive integration strategy in Sec.~\ref{sec:4-6}. 
For further details, including results of targeted attacks, performance under the $\ell_2$ norm, comparative analyses across different surrogate models, comparisons with expanded baseline and more ablation studies, please refer to Appendix \ref{app:d}.

\vspace{-0.5ex}
\subsection{Experimental Settings}\label{sec:4-1}

\textbf{CIFAR-10~\cite{krizhevsky2009learning}.} We adopt all the $10,000$ test images for evaluations, which are in $[0,1]$. We consider 3 black-box target models: ResNet-50 \cite{he2016deep}, DenseNet-121 \cite{huang2017densely}, and SENet-18 \cite{hu2018squeeze}.
We adopt a Wide ResNet model (WRN-34-10) \cite{zagoruyko2016wide} as the surrogate model.
For BO and P-BO, we set an initial dataset $\mathcal{D}$ containing $S=10$ randomly sampled points, and use the Matern-5/2 kernel. The loss function $f$ is the CW loss~\cite{Carlini2016} since it performs better than the cross-entropy loss on CIFAR-10. Following \citet{ru2019bayesopt}, we set the query budget $N=1000$. The perturbation size is $\epsilon=\frac{8}{255}$ under the $\ell_\infty$ norm. As in previous work~\cite{andriushchenko2020square,dong2022query}, we adopt the attack success rate (ASR), the mean and median number of queries of successful attacks to evaluate the performance.

\textbf{ImageNet~\cite{russakovsky2015imagenet}.} We choose $1,000$ images randomly from the ILSVRC 2012 validation set to perform evaluations. Those images are normalized to $[0,1]$. 
The black-box target models include Inception-v3~\cite{szegedy2016rethinking}, MobileNet-v2~\cite{sandler2018mobilenetv2}, and ViT-B/16~\cite{dosovitskiy2020image}.
We use the ResNet-152~\cite{he2016identity} as the surrogate model to provide the prior information. Similar to CIFAR-10, we set $S=10$, $N=1000$, and use the Matern-5/2 kernel. The perturbation size under $\ell_\infty$ norm is $\epsilon=\frac{8}{255}$. When employing dimensionality reduction of the search space, we use $56\times56\times3$, where the original dimensionality is $224\times224\times3$. The subscript ``D'' denotes the methods with dimensionality reduction.

\textbf{MS-COCO~\cite{lin2014microsoft}.} We select $120$ image-caption pairs randomly from MS-COCO for the image captioning task. The images are normalized to $[0,1]$. We consider three black-box VLMs: InstructBLIP~\cite{Dai2023InstructBLIP}, mPLUG-Owl~\cite{ye2023mplug}, and VPGTrans~\cite{zhang2023vpgtrans}. We use MiniGPT-4~\cite{zhu2023minigpt} as the surrogate model. The experimental settings are the same as ImageNet except $\epsilon=\frac{16}{255}$. 
We conduct untargeted attacks on a selected word or phrase describing the main object of each image-caption pair, utilizing the prompt, ``Write a caption for this image''. 
We use log-likelihood between generated caption and selected word as $y_t$ to update the model. 
A successful attack is reported if the generated caption does not contain the specified word or similar words (measured by CLIP-score~\cite{radford2021learning} greater than 0.95).

\vspace{-0.5ex}
\subsection{Experimental Results on CIFAR-10}\label{sec:4-2}
\vspace{-0.3ex}

We compare P-BO with the fixed function prior $\lambda=1$ and adaptive coefficient $\lambda^*$ with various baselines -- NES \cite{ilyas2018black}, Bandits\textsubscript{T} \cite{ilyas2018prior}, $\mathcal{N}$ATTACK \cite{li2019nattack}, SignHunter \cite{al2019sign}, Square attack \cite{andriushchenko2020square}, NPAttack \cite{bai2023query}, RGF \cite{cheng2019improving}, P-RGF \cite{dong2022query}, and BO (i.e., GP-BO in \citet{ru2019bayesopt}). 
For all methods, we restrict the maximum number of queries for each image as $1,000$.
We report a successful attack if a method generates an adversarial example within $1,000$ queries and the size of perturbation is smaller than the budget (i.e., $\epsilon=\frac{8}{255}$).


Table \ref{tab:cifar-l8} shows the results, where we report the attack success rate and the average/median number of queries needed to successfully generate an  adversarial example. Note that for BO and P-BO, the query count includes $S=10$ random queries constructing the initial dataset $\mathcal{D}$. 
We have the following observations.
First, compared with the state-of-the-art attacks, the proposed P-BO generally leads to higher attack success rates and requires much fewer queries.
While most attack methods can achieve success rates above 90\%, only P-BO achieves an absolute 100\% attack success rate. Moreover, P-BO requires less than 20 queries, significantly improving the efficiency. 
Second, the function prior provides useful prior information for black-box attacks since P-BO outperforms the vanilla BO.
Third, P-BO outperforms P-RGF, demonstrating the effectiveness of leveraging the surrogate model as a function prior rather than the gradient prior.
Fourth, the use of the adaptive coefficient $\lambda^*$ in P-BO enhances the attack success rates and reduces the average number of queries compared with $\lambda=1$, highlighting the effectiveness of using an adaptive coefficient derived by the proposed adaptive integration strategy. 

\begin{table*}
\vspace{-1ex}
  \caption{The experimental results of black-box attacks against InstructBLIP, mPLUG-Owl, and VPGTrans under the $\ell_\infty$ norm on MS-COCO. We report the attack success rate (ASR) and the average/median number of queries (AVG. Q/MED. Q) needed to generate an adversarial example over successful attacks. We mark the best results in \textbf{bold}.}
  \label{tab:VLM-l8}
  \small
  \centering
  \begin{tabular}{l|ccc|ccc|ccc}
    \hline
    \multirow{2}{*}{Method} & \multicolumn{3}{c|}{InstructBLIP} & \multicolumn{3}{c|}{mPLUG-Owl} & \multicolumn{3}{c}{VPGTrans}\\
    \cline{2-10}
    & ASR & AVG. Q & MED. Q & ASR & AVG. Q & MED. Q & ASR & AVG. Q & MED. Q \\
    \hline
    RGF\textsubscript{D}~\cite{cheng2019improving} & 35.0\% & 315 & 292 & 46.7\% & 312 & 317 & 28.3\% & 248 & 113 \\
    P-RGF\textsubscript{D}~\cite{dong2022query} & 54.2\% & 358 & 306 & 34.2\% & 274 & 91 & 39.2\% & 296 & 170 \\
    BO\textsubscript{D}~\cite{ru2019bayesopt} & 68.3\% & 96 & 25 & 79.2\% & 58 & 23 & 45.8\% & 84 & 35 \\
    P-BO\textsubscript{D} ($\lambda=1$, \textbf{ours}) & \bf95.8\% & 16 & \bf11 & 83.3\% & \bf24 & \bf14 & 91.7\% & \bf21 & \bf12 \\
    P-BO\textsubscript{D} ($\lambda^*$, \textbf{ours}) & \bf95.8\% & \bf13 & \bf11 & \bf90.8\% & 27 & 16 & \bf96.7\% & 24 & \bf12 \\
    \hline
  \end{tabular}
  \vspace{-1.5ex}
\end{table*}

\begin{table*}
  \caption{The experimental results of black-box attacks against defense models under the $\ell_\infty$ norm on CIFAR-10. We report the attack success rate (ASR) and the average/median number of queries (AVG. Q/MED. Q) needed to generate an adversarial example over successful attacks. We mark the best results in \textbf{bold}.}
  \label{tab:at-l8}
  \small
  \setlength{\tabcolsep}{4.5pt}
  \centering
  \begin{tabular}{l|ccc|ccc|ccc}
    \hline
    \multirow{2}{*}{Method} & \multicolumn{3}{c|}{\citet{rice2020overfitting}} & \multicolumn{3}{c|}{\citet{zhang2019theoretically}} & \multicolumn{3}{c}{\citet{rebuffi2021fixing}}\\
    \cline{2-10}
    & ASR & AVG. Q & MED. Q & ASR & AVG. Q & MED. Q & ASR & AVG. Q & MED. Q \\
    \hline
    NES~\cite{ilyas2018black} & 9.2\% & 324 & 255 & 9.5\% & 321 & 281 & 8.3\% & 328 & 255 \\
    $\mathcal{N}$ATTACK~\cite{li2019nattack} & 18.7\% & 369 & 306 & 19.4\% & 357 & 306 & 15.7\% & 315 & 255 \\
    Square~\cite{andriushchenko2020square} & 13.2\% & 340 & 185 & 15.0\% & 319 & 181 & 11.6\% & 269 & 126 \\
    RGF~\cite{cheng2019improving} & 9.3\% & 268 & 215 & 9.2\% & 259 & 212 & 8.4\% & 281 & 221 \\
    P-RGF~\cite{dong2022query} & 23.4\% & 38 & 33 & 22.8\% & 40 & 31 & 19.1\% & 34 & 31 \\
    BO~\cite{ru2019bayesopt} & 26.1\% & 232 & 129 & 23.7\% & 354 & 288 & 20.4\% & 180 & 112 \\
    P-BO ($\lambda=1$, \textbf{ours}) & \bf36.2\% & 35 & \bf11 & 32.2\% & 38 & \bf11 & 25.9\% & \bf22 & \bf11 \\
    P-BO ($\lambda^*$, \textbf{ours}) & \bf36.2\% & \bf27 & \bf11 & \bf33.9\% & \bf31 & \bf11 & \bf27.2\% & \bf22 & \bf11 \\
    \hline
  \end{tabular}
  \vspace{-1.5ex}
\end{table*}

\vspace{-0.5ex}
\subsection{Experimental Results on ImageNet}\label{sec:4-3}
\vspace{-0.3ex}

Similar to the experiments on CIFAR-10, we also compare the performance of P-BO with these baselines on ImageNet.
We also restrict the maximum number of queries for each image to be $1,000$. 
For some methods including Bandits, RGF, P-RGF, BO, and P-BO, we incorporate the data-dependent prior \cite{ilyas2018prior} into these methods by employing dimensionality reduction of the search space for comparison (which are denoted by adding a subscript ``D'').

The black-box attack performance of those methods is presented in Table \ref{tab:imagenet-l8}. It can be seen that our P-BO achieves the highest ASR with the lowest average query count, outperforming the existing methods. 
Furthermore, within the P-BO algorithm, fixing the adaptive integration coefficient $\lambda=1$ results in inferior performance compared with the baseline BO algorithm without incorporating priors. As discussed in Sec.~\ref{sec:3-2}, this phenomenon is attributed to the condition $\|f-f'\|_k > \|f\|_k$ in Theorem \ref{thm:1}, which may be due to the lower similarity between models on the ImageNet dataset. This underscores the importance of adaptive tuning of the integration coefficient $\lambda$: when $\lambda$ is adaptive, the P-BO algorithm exhibits higher success rates and significantly reduces queries, demonstrating stability in dealing with varying levels of prior's effectiveness.
Additionally, the results also validate that the data-dependent prior is orthogonal to the proposed function prior, since integrating the data-dependent prior leads to better results.

\begin{figure}[t]
    \centering   \includegraphics[width=0.99\linewidth]    {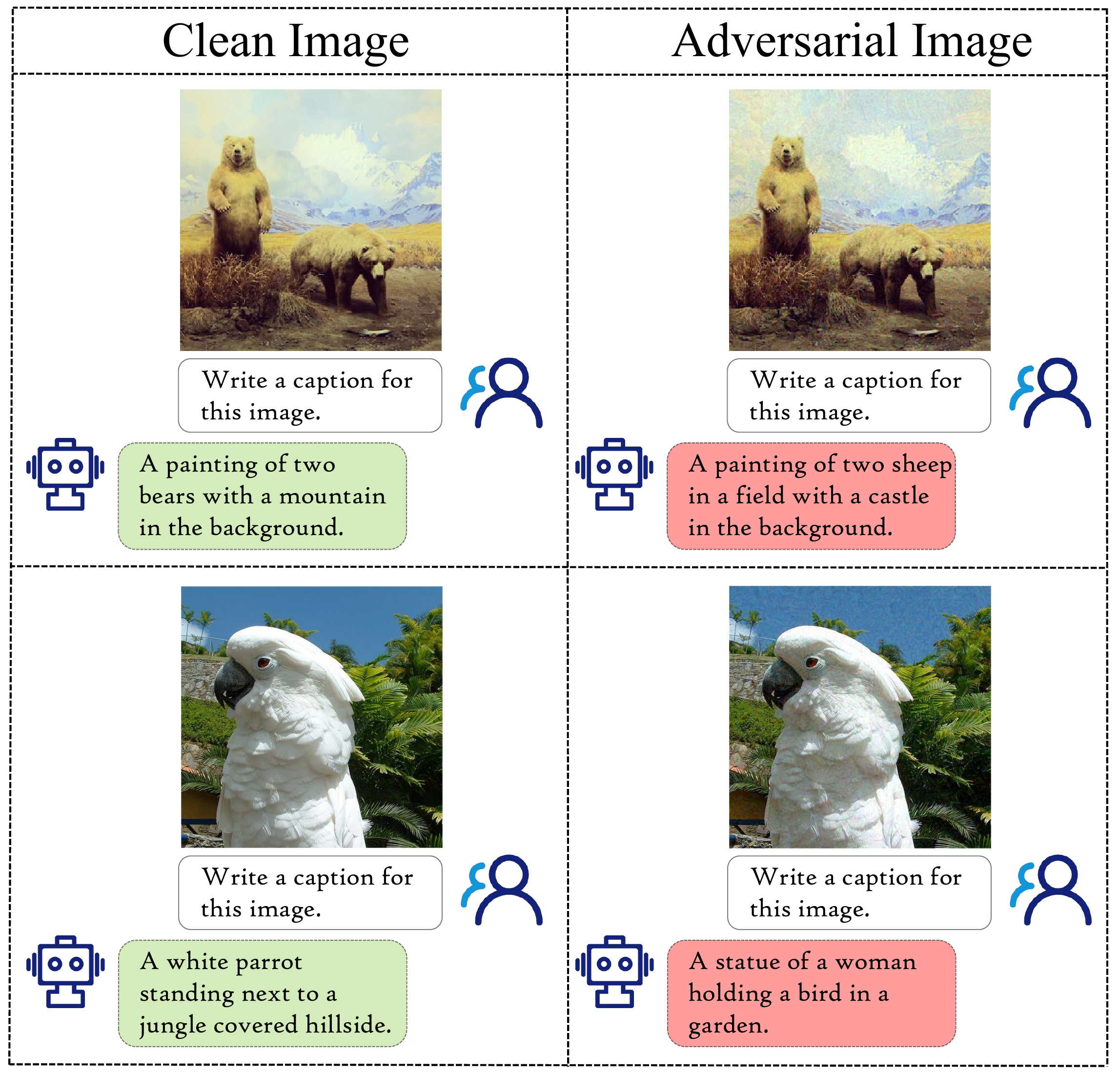}
    \vspace{-2ex}
    \caption{We show two adversarial examples against InstructBLIP. They mislead the VLM to output wrong descriptions.}    \label{fig:fig2}
    \vspace{-3ex}
\end{figure}

\vspace{-0.5ex}
\subsection{Experimental Results on Vision-Language Models}\label{sec:4-4}
\vspace{-0.3ex}

Large Vision-Language Models (VLMs) have achieved unprecedented performance in response generation, especially with visual inputs, enabling more creative and adaptable interaction. Nonetheless, multimodal generation exacerbates safety concerns, since adversaries may successfully evade the entire system by subtly manipulating the most vulnerable modality (e.g., vision) \cite{dong2023robust,zhao2023evaluating}. In this section, we present the results of black-box adversarial attacks against VLMs.
We compare the performance of P-BO with three strong baselines: RGF~\cite{cheng2019improving}, P-RGF~\cite{dong2022query}, and BO~\cite{ru2019bayesopt}. These methods adopt the dimensionality reduction technique. For all methods, we limit the maximum query count for each image to $1,000$.

Table \ref{tab:VLM-l8} shows the attack results against InstructBLIP~\cite{Dai2023InstructBLIP}, mPLUG-Owl~\cite{ye2023mplug}, and 
VPGTrans \cite{zhang2023vpgtrans}. Our P-BO algorithm achieves at least 90\% success rates for all target VLMs, with a low average query count, demonstrating high query efficiency. Although P-BO ($\lambda=1$) has a slightly lower average query count than P-BO ($\lambda^*$), there is a noticeable gap in the attack success rate. Fig.~\ref{fig:fig2}  shows two adversarial examples generated by P-BO, which mislead InstructBLIP to incorrectly describe the image contents. 
These results showcase the generalizability of P-BO in conducting black-box attacks against large VLMs, emphasizing the need for a more thorough examination of their potential security flaws before deployment.

\subsection{Experimental Results on Defense Models}\label{sec:4-5}
We include three defense models \cite{rice2020overfitting,zhang2019theoretically,rebuffi2021fixing} on CIFAR-10 as the targets to perform black-box adversarial attacks.
They are all based on adversarial training. 
The experimental settings are the same as those in Sec.~\ref{sec:4-2}. We adopt \citet{zhang2019theoretically} as the surrogate model when attacking the others, while adopting \citet{rice2020overfitting} as the surrogate model when attacking \citet{zhang2019theoretically}, since a normally trained model can be hardly useful for attacking defenses~\cite{dong2020benchmarking}.

We compare the performance of P-BO with six baselines, including NES~\cite{ilyas2018black}, $\mathcal{N}$ATTACK~\cite{li2019nattack}, RGF~\cite{cheng2019improving}, P-RGF~\cite{dong2022query}, Square attack~\cite{andriushchenko2020square}, and BO \cite{ru2019bayesopt}. The attack results are presented in Table \ref{tab:at-l8}. Similar to the results on the normal models, the proposed P-BO method can achieve higher success rates and require much less queries than the other baselines. The experiments on adversarial defenses consistently validate the effectiveness of our proposed methods.

\begin{figure*}[t]
\centering
\includegraphics[width=0.95\textwidth]
{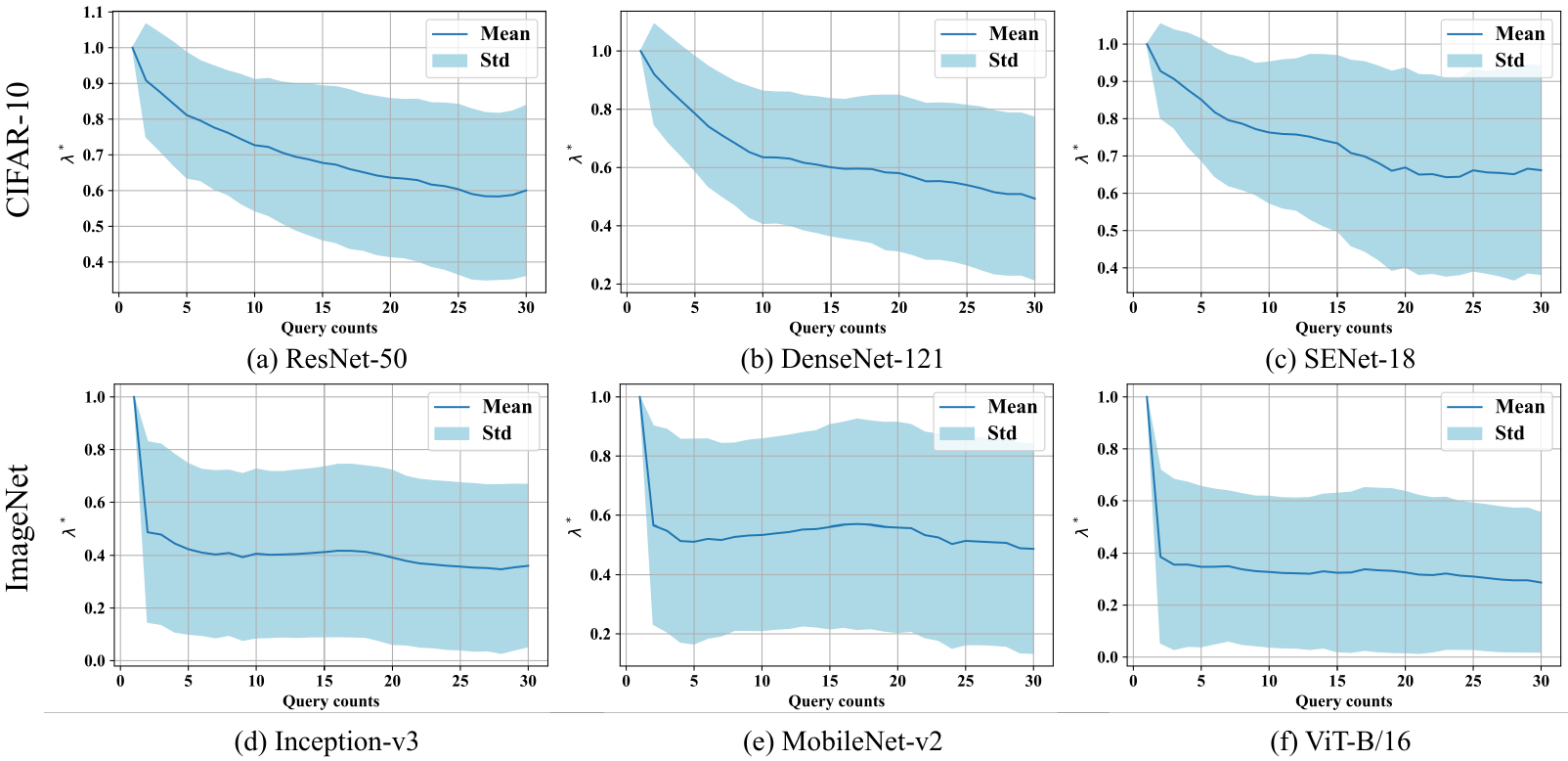}
\vspace{-1ex}
\caption{The mean and standard deviation of $\lambda^*$ over the first 30 iterations of P-BO applied to different target models on CIFAR-10 and ImageNet. $\lambda^*$ on ImageNet is substantially lower than that on CIFAR-10. This implies a lower similarity between different ImageNet models, and thus the function prior might be less useful. }
\label{fig:fig3}
\vspace{-1ex}
\end{figure*}

\subsection{Performance of Adaptive Integration Strategy}\label{sec:4-6}
We conduct experiments to investigate the trends of $\lambda^*$ across different target models on CIFAR-10 and ImageNet. Specifically, we show the mean and standard deviation of $\lambda^*$ over the first 30 iterations of P-BO. Note that $\lambda^*$ is initialized as 1 in the first iteration.
As shown in Fig. \ref{fig:fig3}, 
$\lambda^*$ on ImageNet is significantly smaller than that on CIFAR-10, e.g., at the 5-th iteration, $\lambda^*\approx 0.8$ for various target models on CIFAR-10, whereas $\lambda^* \approx 0.4$ on ImageNet.
This implies a lower similarity between different ImageNet models, and thus the function prior might be less useful. 
This can also explain why P-BO with $\lambda=1$ has inferior performance than BO since $\lambda$ is too large in this case.
However, the adaptive integration coefficient $\lambda^*$ in P-BO can be dynamically decreased to ensure that its performance remains unaffected by a useless prior, leading to consistent improvements over BO.
Besides, $\lambda^*$ for ViT-B/16 is lower than that for Inception-v3 and MobileNet-v2 on ImageNet. This discrepancy is caused by different architectures of ViT-B/16 (transformer) and the surrogate model (CNN).

\section{Conclusion}

In this paper, we propose a Prior-guided Bayesian Optimization (P-BO) method for more query-efficient black-box adversarial attacks. P-BO models the attack objective function with a Gaussian process, whose mean function is initialized by a function prior, i.e., the surrogate model's loss function. After updating the posterior distribution given the observations, the next query point is chosen by maximizing an acquisition function. We analyze the regret bound of P-BO, which is proportional to the RKHS norm between the objective function and the function prior. To avoid performance degradation in case of a bad prior, we further propose an adaptive integration strategy which automatically adjusts a coefficient on the function prior. Extensive experiments consistently demonstrate the effectiveness of P-BO in reducing the number of queries and improving attack success rates.

\vspace{-1ex}
\section*{Impact Statement}
A potential negative societal impact of our work is that malicious adversaries may adopt our method to efficiently query actual victim systems for generating adversarial samples in the real world, which can cause severe security/safety consequences for real-world applications. Thus it is imperative to develop more robust models against our attack, which we leave to future work.

\section*{Acknowledgements}

This work was supported by the National Natural Science Foundation of China (Nos. 62276149, 92370124, 62350080, 92248303, U2341228, 62061136001, 62076147, 12288201), 
the National Key Research and Development Plan (No. 2018YFA0306702), BNRist (BNR2022RC01006), 
Tsinghua Institute for Guo Qiang, and the High Performance Computing Center, Tsinghua University. Y. Dong was also supported by the China National Postdoctoral Program for Innovative Talents. J. Zhu was also supported by the XPlorer Prize.

\bibliography{example_paper}
\bibliographystyle{icml2024}

\newpage
\appendix
\onecolumn



\section{Proofs} \label{app:a}

\subsection{Proof of Theorem~\ref{thm:1}}\label{app:a-1}

\begin{proof}
Let $\bm{x}^*=\argmax_{\bm{x}\in A}f(\bm{x})$ be the global maximum, and $\bm{x}_{t}= \operatorname{argmax}_{\bm{x}\in A}\mu_{t-1}(\bm{x})+\|f-f'\|_k \sigma_{t-1}(\bm{x})$ be the next query point. 
We first prove that for $\forall \bm{x}\in A, t\leq T$, we have
\begin{align}\label{eq:111}
        |f(\bm{x})-\mu_{t-1}(\bm{x})|\leq \|f-f'\|_k \sigma_{t-1}(\bm{x}).
\end{align}

If $f'\equiv0$, due to the reproducing property of RKHS, we have $|f(\bm{x})-\mu_{t-1}(\bm{x})|\leq \sigma_{t-1}(\bm{x}) \|f-\mu_{t-1}\|_k$.
Then we prove that $\|f-\mu_{t-1}\|_k\leq \|f\|_k\ $.
Let $(a_1,\ldots,a_{t-1})^\top =\mathbf{K}_{t-1}^{-1} \mathbf{y}_{t-1}$. Then $\mu_{t-1}=\sum_{i=1}^{t-1} a_i k(\bm{x}_i,\cdot)$. Thus, 
$\langle f, \mu_{t-1}\rangle_k =\langle \mu_{t-1}, \mu_{t-1}\rangle_k=\mathbf{y}_{t-1}^\top \mathbf{K}_{t-1}^{-1} \mathbf{K}_{t-1} \mathbf{K}_{t-1}^{-1} \mathbf{y}_{t-1}=\mathbf{y}_{t-1}^\top \mathbf{K}_{t-1}^{-1} \mathbf{y}_{t-1}$.
So, we have 
\begin{align}
\|f-\mu_{t-1}\|_k^2= \|f\|_k^2\ - 2\langle f, \mu_{t-1}\rangle_k + \|\mu_{t-1}\|_k^2\ = \|f\|_k^2\ - \mathbf{y}_{t-1}^\top \mathbf{K}_{t-1}^{-1} \mathbf{y}_{t-1}. 
\end{align}
Thus $\|f-\mu_{t-1}\|_k\leq \|f\|_k$, and $|f(\bm{x})-\mu_{t-1}(\bm{x})|\leq \|f\|_k \sigma_{t-1}(\bm{x})$.

Next, we consider the general case, where $f' \neq 0$.
Let $\mu_{t-1}'(\bm{x}):=\mu_{t-1}(\bm{x})-f'(\bm{x})$ and $\sigma_{t-1}'(\bm{x}):=\sigma_{t-1}(\bm{x})$, we have
\begin{align}
|(f-f')(\bm{x})-\mu_{t-1}'(\bm{x})|\leq \|f-f'\|_k \sigma_{t-1}'(\bm{x}).
\end{align}
Therefore, substituting $\mu_{t-1}'(\bm{x})$ and $\sigma_{t-1}'(\bm{x})$, we obtain $|f(\bm{x})-\mu_{t-1}(\bm{x})|\leq \|f-f'\|_k \sigma_{t-1}(\bm{x})$.

Then we prove that the instantaneous regret $r_t\leq 2\|f-f'\|_k\sigma_{t-1}(\bm{x}_t)$.
By the UCB acquisition function defined in Eq.~\eqref{eq:ucb} and definition of $\bm{x}_{t}$, we have
\begin{align}
\mu_{t-1}(\bm{x}_{t})+\|f-f'\|_k \sigma_{t-1}(\bm{x}_{t}) \geq \mu_{t-1}(\bm{x}^*)+\|f-f'\|_k \sigma_{t-1}(\bm{x}^*).
\end{align}
By plugging in Eq.~\eqref{eq:111} at ${\bm{x}}^*$:
\begin{align}
|f({\bm{x}}^*)-\mu_{t-1}({\bm{x}}^*)|\leq \|f-f'\|_k \sigma_{t-1}({\bm{x}}^*),
\end{align}
we have that $\mu_{t-1}(\bm{x}^*)+\|f-f'\|_k \sigma_{t-1}(\bm{x}^*)\geq f(\bm{x}^*)$.
So the instantaneous regret has
\begin{align}
r_t=f(\bm{x}^*)-f(\bm{x}_t)
\leq \mu_{t-1}(\bm{x}_{t})+\|f-f'\|_k \sigma_{t-1}(\bm{x}_{t}) -f(\bm{x}_{t}).
\end{align}
Through a recursive application of Eq.~\eqref{eq:111} at $\bm{x}_{t}$:
\begin{align}
|f(\bm{x}_{t})-\mu_{t-1}(\bm{x}_{t})|\leq \|f-f'\|_k \sigma_{t-1}(\bm{x}_{t}),
\end{align}
we arrive at the revelation that 
\begin{align}\label{eq:222}
r_t\leq 2\|f-f'\|_k\sigma_{t-1}(\bm{x}_t).
\end{align}

In the third step, we prove that 
\begin{align}
\sum_{t=1}^T \sigma_{t-1}(\bm{x}_t)^2\leq \frac{2}{\log (1+\sigma^{-2})}\gamma_T,
\end{align}
where $\gamma_T=\frac{1}{2}\max_{\bm{x}_1,\ldots,\bm{x}_T\in A}\log |\mathbf{I}+\sigma^{-2}\mathbf{K}_T|$.

We have $\sigma_{t-1}(\bm{x}_t)^2 = \sigma^2 (\sigma^{-2} \sigma_{t-1}(\bm{x}_t)^2)$.
Since ${x}^2 \leq C \log(1+{x}^2)$ for ${x}\in \left[0, \sigma^{-2}\right]$, $C\geq1$, and $\sigma^{-2} \sigma_{t-1}(\bm{x}_t)^2\leq \sigma^{-2} k(\bm{x}_t, \bm{x}_t) \leq \sigma^{-2}$, $\frac{\sigma^{-2}}{\log (1+\sigma^{-2})} \geq 1$, 
we have
\begin{align}
\sigma^{-2} \sigma_{t-1}(\bm{x}_t)^2 \leq \frac{\sigma^{-2}}{\log (1+\sigma^{-2})} \log (1+\sigma^{-2} \sigma_{t-1}(\bm{x}_t)^2).
\end{align}
Thus,
$\sigma_{t-1}(\bm{x}_t)^2 \leq \frac{1}{\log (1+\sigma^{-2})} \log (1+\sigma^{-2} \sigma_{t-1}(\bm{x}_t)^2)$.
Then we can derive that
\begin{align}\label{eq:333}
\sum_{t=1}^T \sigma_{t-1}(\bm{x}_t)^2\leq \frac{2}{\log (1+\sigma^{-2})} \frac{1}{2} \sum_{t=1}^T \log (1+\sigma^{-2} \sigma_{t-1}(\bm{x}_t)^2).
\end{align}

On the other hand, we also have
\begin{align}
\frac{1}{2}\log |\mathbf{I}+\sigma^{-2}\mathbf{K}_T| = \mathrm{H}(\mathcal{N}(\mathbf{y}'_T, \mathbf{K}_T))-\mathrm{H}(\mathcal{N}(\mathbf{y}_T, \sigma^2\mathbf{I})),
\end{align}
where $\mathrm{H}(\cdot)$ represents the entropy of a distribution.
Hence,
\begin{align}
\frac{1}{2}\log |\mathbf{I}+\sigma^{-2}\mathbf{K}_T| 
&= \mathrm{H}(\mathcal{N}(\mathbf{y}'_T, \mathbf{K}_T))-\mathrm{H}(\mathcal{N}(\mathbf{y}_T, \sigma^2\mathbf{I})) \\
&= \mathrm{H}(\mathcal{N}(\mathbf{y}'_T, \mathbf{K}_T))-\frac{1}{2} \log \left| 2 {\pi} e \sigma^{2} \mathbf{I}\right| \\
&= \mathrm{H}(\mathcal{N}(\mathbf{y}'_{T-1}, \mathbf{K}_{T-1})) + \mathrm{H}(\mathcal{N}(\mathbf{y}'_T | \mathbf{y}'_{T-1})) -\frac{1}{2} \log \left| 2 {\pi} e \sigma^{2} \mathbf{I}\right| \\
&= \mathrm{H}(\mathcal{N}(\mathbf{y}'_{T-1}, \mathbf{K}_{T-1})) + \frac{1}{2} \log \left( 2 {\pi} e (\sigma^{2} + \sigma_{T-1}(x_T)^2)\right) -\frac{1}{2} \log \left| 2 {\pi} e \sigma^{2} \mathbf{I}\right|.
\end{align}
By means of induction, we obtain 
\begin{align}\label{eq:444}
\frac{1}{2}\log |\mathbf{I}+\sigma^{-2}\mathbf{K}_T| = \frac{1}{2} \sum_{t=1}^T \log (1+\sigma^{-2} \sigma_{t-1}(\bm{x}_t)^2).
\end{align}

Substituting Eq.~\eqref{eq:444} into Eq.~\eqref{eq:333}, we will obtain the proof that
\begin{align}\label{eq:555}
\sum_{t=1}^T \sigma_{t-1}(\bm{x}_t)^2\leq \frac{2}{\log (1+\sigma^{-2})}\gamma_T.
\end{align}

Since $R_T=\sum_{t=1}^T r_t$, using Eq.~\eqref{eq:222}, Eq.~\eqref{eq:555} and Cauchy Schwarz inequality, we can derive that
\begin{align}\label{eq:666}
R_T 
&\leq \sqrt{T \sum_{t=1}^T r_t} \\
&\leq 2\|f-f'\|_k \sqrt{T \sum_{t=1}^T \sigma_{t-1}(\bm{x}_t)^2} \\
&\leq \|f-f'\|_k\sqrt{\frac{8}{\log (1+\sigma^{-2})}T\gamma_T}.
\end{align}
\end{proof}

\subsection{Proof of Proposition~\ref{thm:2}}\label{app:a-2}

\begin{proof}
We provide a proof based on that of \citet{kanagawa2018gaussian}. Let 
\begin{align}
\mathcal{F}_k := \left\{\sum_{t=1}^T \beta_t k(\bm{x}_t, \cdot)|  \forall\beta_1,\dots, \beta_T \in \mathbb{R} \right\},
\end{align}
\begin{align}
\mathcal{H}_k := \left\{ h \in \mathcal{H}| h(\bm{x}_t) = y_t,\forall t\leq T \right\}.
\end{align}
We first prove that $\{\mu_T\} = \mathcal{F}_k \cap \mathcal{H}_k$.
Notice that $\mu_T=\sum_{t=1}^T a_t k(\bm{x}_t,\cdot)$, where $(a_1,\ldots,a_T)^\top =\mathbf{K}_T^{-1} \mathbf{y}_T$, therefore, $\mu_T \in \mathcal{F}_k$. Also, $\mu_T(\bm{x}_t) = y_t$, then $\mu_T \in \mathcal{H}_k$. Hence, $\mu_T \in \mathcal{F}_k \cap \mathcal{H}_k$.

Next, we will prove that if $f \in \mathcal{F}_k \cap \mathcal{H}_k$, then $f=\mu_T$.
Assume that $f \in \mathcal{F}_k \cap \mathcal{H}_k$. let $f = \sum_{t=1}^T \gamma_t k(\bm{x}_t, \cdot)$, $\gamma_t \in \mathbb{R}$. Then
\begin{align}
\mu_T - f = \sum_{t=1}^T (a_t-\gamma_t) k(\bm{x}_t, \cdot) \in \mathcal{F}_k.
\end{align}
Also, since $f \in \mathcal{H}_k$, then $f(\bm{x}_t) = y_t$, we have
\begin{align}
\left<\mu_T-f, k(\bm{x}_t, \cdot) \right>_{k} = \mu_T(\bm{x}_t) - f(\bm{x}_t) = y_t - y_t = 0.
\end{align}
so $\mu_T-f \perp \mathcal{F}_k$, which implies $\mu_T-f \in \mathcal{F}_k \cap \mathcal{F}_k^\perp = \{ 0 \}$. Thus, $\{ \mu_T \} = \mathcal{F}_k \cap \mathcal{H}_k$.

Then we prove that $\mu_T = \argmin_{h\in\mathcal{H}_k} \|h\|_k$. Since $\mathcal{H}_k$ is convex and closed, there exists an solution $h^* \in \mathcal{H}_k$ such that $h^* = \argmin_{h\in\mathcal{H}_k} \|h\|_k$. For $\forall g \perp \mathcal{F}_k$, we have
\begin{align}
\left< h^* + g,  k(\bm{x}_t, \cdot) \right>_{k} = \left< h^*,  k(\bm{x}_t, \cdot) \right>_{k} = h^*(\bm{x}_t) = y_t.
\end{align}
So $h^* + g \in \mathcal{H}_k$. Since $\| h^* \|_{k} \leq \| h^* + g \|_{k}$ and $\mathcal{F}_k$ is closed, we have $h^* \in (\mathcal{F}_k^\perp)^\perp = \mathcal{F}_k$. So $h^* \in \mathcal{F}_k \cap \mathcal{H}_k$, and $\mathcal{F}_k \cap \mathcal{H}_k = \{ \mu_T \}$, thus, $h^* = \mu_T$. Since $\langle k(\bm{x}_{t_i},\cdot), k(\bm{x}_{t_j},\cdot)\rangle_k=k(\bm{x}_{t_i},\bm{x}_{t_j})$ and $\mu_T=\sum_{t=1}^T a_t k(\bm{x}_t,\cdot)$, we have 
\begin{align}
\|\mu_T\|_k^2=\langle \mu_T, \mu_T\rangle_k=\mathbf{y}_T^\top \mathbf{K}_T^{-1} \mathbf{K}_T \mathbf{K}_T^{-1} \mathbf{y}_T=\mathbf{y}_T^\top \mathbf{K}_T^{-1} \mathbf{y}_T.
\end{align}
\end{proof}

\section{Related Work} \label{app:b}

\textbf{Query-based attack methods.} 
Query-based methods generate adversarial examples by leveraging the query feedback of the black-box model. 
ZOO~\cite{Chen2017ZOO} uses symmetric difference to estimate gradient for every pixel.
NES \cite{ilyas2018black} utilizes natural evolution strategy to estimate gradients.
Bandits \cite{ilyas2018prior} improves the NES method by incorporating data and temporal priors into the gradient estimation.
$\mathcal{N}$ATTACK \cite{li2019nattack} introduces a gradient estimation framework to improve the attack success over defensive models.
RGF \cite{cheng2019improving} utilizes random gradient-free method to estimate gradients.
SimBA \cite{guo2019simple} adapts a greedy strategy to update the query samples.
Square Attack \cite{andriushchenko2020square} introduces highly-efficient greedy random search for black-box adversarial attack.
SignHunter \cite{al2019sign} adapts the gradient sign rather than the gradient as the search direction.
NPAttack \cite{bai2023query} explore the distribution of adversarial examples around benign inputs with the help of image structure information characterized by a Neural Process.
BayesOpt \cite{ru2019bayesopt} and similar methods \cite{huang2020corrattack,li2023bayesian} utilize Bayesian optimization for black-box attacks.
AdvFlow \cite{mohaghegh2020advflow} approximates the adversarial distribution with the clean data distribution
PPBA \cite{li2020projection} shrinks the solution space of possible adversarial inputs to those which contain low- frequency perturbations.
PRFA \cite{liang2022parallel} proposes to parallelly attack multiple rectangles for better efficiency.
Additionally, several studies apply BO-based black-box attacks to various specific application scenarios, including graph classification, natural language processing, and protein classification.
\citet{wan2021adversarial} introduce a novel BO-based attack method tailored for graph classification models. They provide valuable insights into the relationship between changes in graph topology and model robustness.
\citet{lee2022query} put forward a query-efficient black-box attack leveraging BO, which dynamically determines crucial positions using an automatic relevance determination (ARD) categorical kernel.
In a subsequent work, \citet{lee2023query} propose innovative query-efficient black-box red teaming methods based on BO, specifically targeting large-scale generative models.
The main limitation of these methods is that they inevitably require a tremendous number of queries to perform a successful attack, leading to a low attack success rate given a limited query budget. 

\textbf{Combination-based attack methods.} 
Combination-based methods \cite{cheng2019improving,guo2019subspace,du2019query,yang2020learning,ma2020switching,huang2020black,suya2020hybrid,yatsura2021meta,lord2022attacking,feng2022boosting,dong2022query,yin2023generalizable} integrate transfer-based attacks with query-based attacks to achieve high attack success rate and high query efficiency simultaneously. 
\citet{cheng2019improving,dong2022query} propose P-RGF, utilizing the gradient of the surrogate model as a transfer-based prior to obtain a more accurate gradient estimate. 
\citet{guo2019subspace,ma2020switching,yang2020learning} restrict the search space
spanned by the surrogate gradients.
Subspace attack \cite{guo2019subspace} regards surrogates’ transfer-prior as subspaces to reduce search space of random vectors with gradients.
LeBA \cite{guo2019subspace} proposes to learn the victim’s estimated gradients via high-order computation graph.
Although these methods are effective in expediting convergence and reducing the number of queries, the surrogate gradient is localized and can be misleading, limiting their effectiveness. 
Besides, other methods learn a generalizable model-based prior based on the surrogate model \cite{du2019query,huang2020black,yin2023generalizable}. 
Meta attack \cite{du2019query} adopts meta-learning to approximate the victim.
TREMBA \cite{huang2020black} treats the projection from a low-dimensional space to the original space as the prior, such that the perturbation could be search in the low-dimensional space.
MCG \cite{yin2023generalizable} trains a meta generator to produce perturbations conditioned on benign examples.
But these methods require an additional dataset to train the attack generator, which is not applicable when the data is scarce.

\textbf{Bayesian optimization with prior.} 
In Bayesian optimization, the incorporation of prior information can be broadly classified into three types~\cite{hvarfner2022pi}. 
The first type of prior information pertains to the distribution of the optimal solution's location. \citet{snoek2014input,ramachandran2020incorporating} enhance exploration in local regions and suppress exploration in unimportant areas by warping the input space. In contrast, \citet{hvarfner2022pi,li2020incorporating,souza2021bayesian} introduce a prior distribution on the location of the optimal solution and compute the posterior distribution of the optimal solution's location given observed data. Such prior information is typically proposed by experienced practitioners or domain experts, and its form is often straightforward.
The second type of prior information concerns the structure of the objective function. For instance, \citet{li2019accelerating} utilize the monotonic trend of the independent variables to model. These methods necessitate a concrete understanding or assumption about the nature of the objective function.
The third type of prior information is derived from datasets obtained from similar unknown objective functions, and the methods utilizing such prior information are referred to as transfer learning in Bayesian optimization. There are generally two approaches~\cite{tighineanu2022transfer,shen2023proxybo}: the first involves jointly modeling datasets corresponding to both prior information and the data obtained from the target function~\cite{swersky2013multi,yogatama2014efficient,poloczek2017multi}, while the second involves separately modeling the dataset corresponding to the prior information to aid in modeling the target function~\cite{golovin2017google,feurer2018scalable,wistuba2018scalable}. The latter approach often involves fitting a Gaussian process on the prior dataset, treating the predictive distribution of the Gaussian process as a prior with uncertainty. 
Specifically, \citet{feurer2018scalable,wistuba2018scalable} weightedly average the predicted mean and variance of the Gaussian process fitted on the prior dataset with those fitted on the objective function dataset, with heuristic weights lacking theoretical analysis. \citet{golovin2017google} employ the difference between the fitted Gaussian process predicted mean on the target function dataset and that on the prior dataset for fusion but do not incorporate adaptive weights.
According to our analysis of the regret upper bound, non-adaptive fusion might degrade algorithm performance. Additionally, this method's setting for predictive variance is not suitable for the deterministic function prior described in our black-box attack scenario.
None of the aforementioned algorithms analyze the regret upper bound of Bayesian optimization algorithms, and their settings differ from the scenario where a deterministic function prior is directly provided. Moreover, these methods have not been applied to black-box attacks.

\section{A Case of Random Linear Function} \label{app:c}

In light of Theorem~\ref{thm:1}, it is evident that although convergence is guaranteed when modeling with $\mathrm{GP}(f', k)$, achieving improved algorithmic performance is desirable to have a small value for $\|f-f'\|_k$, at least satisfying $\|f-f'\|_k \leq \|f\|_k$. While this naturally holds when $f'$ closely approximates $f$, in the case of functions defined in high-dimensional spaces, this condition is quite stringent. As follows, we provide a natural counterexample, demonstrating the challenge in meeting this condition, when the functions are defined in high-dimensional spaces, exhibiting a tendency towards orthogonality.
Let $f$ and $f'$ both be linear functions, where $f(\bm{x})=\bm{w}^\top \bm{x}$, $f'(\bm{x})=\bm{w}'^\top \bm{x}$. Consequently, $(f-f')(\bm{x})=(\bm{w}-\bm{w}')^\top \bm{x}$. Assuming the kernel function $k$ is isotropic and stationary, i.e., $k(\bm{x},\bm{x}')$ depends only on $\|\bm{x}-\bm{x}\|_2$, in accordance with the absolute homogeneity of norms, $\|f-f'\|_k\leq \|f\|_k$ is equivalent to $\|\bm{w}-\bm{w}'\|_2\leq \|\bm{w}\|_2$. Assuming $\|\bm{w}\|_2=\|\bm{w}'\|_2=1$, this requires $\bm{w}^\top \bm{w}'\geq \frac{1}{2}$.
Considering that if $\bm{w}, \bm{w}'$ are uniformly sampled from the unit hypersphere in $\mathbb{R}^d$, the expected value $\E[(\bm{w}^\top \bm{w}')^2]=\frac{1}{d}$. It becomes evident that the probability of $\bm{w}^\top \bm{w}'\geq \frac{1}{2}$ is indeed small.
This implies that when the target and prior functions are not closely aligned, a direct $\mathrm{GP}(f', k)$ modeling approach may lead to $\|f-f'\|_k > \|f\|_k$. Such a scenario could result in the prior-guided Bayesian optimization algorithm's performance degradation compared to the approach in which prior information is not incorporated.

\section{Supplementary Experimental Results} \label{app:d}

\subsection{More Experimental Details}

We conduct the experiments on NVIDIA 2080 Ti (for CIFAR-10 and ImageNet) and A100 (for VLMs) GPUs. The source code of P-BO is submitted in the supplementary material and will be released after the review process. For the baseline attacks, we adopt the official implementations of NES, Bandits, $\mathcal{N}$ATTACK, SignHunter, Square attack, and NPAttack. For RGF and P-RGF, we adopt a better implementation: we make the random directions $\{\bm{u}_i\}_{i=1}^q$ orthogonal following \citet{cheng2021convergence}; we set $q=5$, $\sigma=0.05$, $lr=0.1$ on CIFAR-10, and $\sigma=0.5$, $lr=0.1$ on ImageNet and MS-COCO.\footnote{In our implementation of RGF, P-RGF, BO and P-BO, instead of directly optimizing w.r.t. $\bm{x}$, we reparametrize $\bm{x}$ with $\bm{x}=\bm{x}^{nat} + \epsilon\cdot \bm{\delta}$ and optimize w.r.t. $\bm{\delta}$ where $\bm{\delta}\in [-1, 1]^D$. Therefore, the mentioned $\sigma$ and learning rate $lr$ in RGF and P-RGF are under the context of the search space $[-1, 1]^D$. \label{footnote:hp-scale}} We adopt the gradient averaging method in P-RGF due to its better performance \cite{dong2022query}.
We implement highly efficient BO and P-BO algorithms in PyTorch by referring to the implementation of \citet{ru2019bayesopt} based on SciPy (and NumPy). One main difference is that when optimizing the acquisition function, we use the PGD algorithm with learning rate 0.05 and 50 iterations of optimization instead of the L-BFGS method, and we can try multiple random starts in parallel. We initialize the length scale in the Matern-5/2 kernel as $\sqrt{D}$.\footnote{The mentioned learning rate and length scale in BO and P-BO are under the context of the search space $[-1, 1]^D$.}
The balancing coefficient $\beta$ is set to 3. Our implementation of BO has similar performance to \citet{ru2019bayesopt}.

\subsection{Targeted Attacks on CIFAR-10}

In this section, we perform black-box targeted adversarial attacks against three CIFAR-10 models, including ResNet-50~\cite{he2016deep}, DenseNet-121~\cite{huang2017densely}, and SENet-18~\cite{hu2018squeeze}. We select $100$ correctly classified images and target at all other $9$ classes, leading to $900$ trials.
We compare the performance of P-BO with three strong baselines --- RGF~\cite{cheng2019improving}, P-RGF~\cite{dong2022query}, and BO \cite{ru2019bayesopt}. 
For all methods, we restrict the maximum number of queries for each image to be $1$,$000$, and the experimental settings are aligned with those outlined in Sec. \ref{sec:4-2}.

Table \ref{tab:cifar-l8-target} shows the results, where we report the success rate of black-box targeted attacks and the average/median number of queries needed to generate an adversarial example over successful attacks.
Compared with the state-of-the-art attacks, the proposed method P-BO generally leads to higher attack success rates and requires much fewer queries.
The P-BO algorithm demonstrates a notable improvement over conventional BO algorithms, attaining remarkably high success rates with only a few dozen average query counts, showcasing the effectiveness and practicality of P-BO.
The adaptive coefficient $\lambda^*$ in P-BO, in comparison to a fixed coefficient $\lambda=1$, significantly enhances the attack success rate and reduces the average query count, underscoring the efficacy of utilizing adaptive fusion weights. 
Additionally, the improvement of P-BO over the baseline BO algorithm is more pronounced than the enhancement observed in P-RGF over RGF, indicating that the P-BO algorithm efficiently utilizes useful function prior information to a considerable extent.

\begin{table*}
  \caption{The experimental results of black-box targeted attacks against DenseNet-121, ResNet-50, and SENet-18 under the $\ell_\infty$ norm on CIFAR-10. We report the attack success rate (ASR) and the average/median number of queries (AVG. Q/MED. Q) needed to generate an adversarial example over successful attacks. We mark the best results in \textbf{bold}.}
  \label{tab:cifar-l8-target}
  \centering\small
  \begin{tabular}{l|ccc|ccc|ccc}
    \hline
    \multirow{2}{*}{Methods} & \multicolumn{3}{c|}{ResNet-50} & \multicolumn{3}{c|}{DenseNet-121} & \multicolumn{3}{c}{SENet-18}\\
    \cline{2-10}
    & ASR & AVG. Q & MED. Q & ASR & AVG. Q & MED. Q & ASR & AVG. Q & MED. Q \\
    \hline
    RGF~\cite{cheng2019improving} & 47.6\% & 389 & 325 & 39.7\% & 392 & 322 & 44.1\% & 363 & 316 \\
    P-RGF~\cite{dong2022query} & 72.2\% & 184 & 70 & 60.6\% & 219 & 102 & 71.8\% & 167 & 60 \\
    BO~\cite{ru2019bayesopt} & 82.7\% & 223 & 140 & 74.1\% & 244 & 155 & 81.4\% & 225 & 135 \\
    P-BO\textsubscript{D} ($\lambda=1$, \textbf{ours}) & 96.9\% & 42 & \bf21 & 91.2\% & 97 & \bf26 & 96.8\% & 45 & \bf16 \\
    P-BO\textsubscript{D} ($\lambda^*$, \textbf{ours}) & \bf98.7\% & \bf34 & \bf21 & \bf97.1\% & \bf55 & 28 & \bf99.2\% & \bf33 & 21 \\
    \hline
  \end{tabular}
\end{table*}

\subsection{Experimental Results under the $\ell_2$ Norm on CIFAR-10}

We further conduct experiments under the $\ell_2$ norm on CIFAR-10. 
The experimental settings are the same as those in Sec. Sec.~\ref{sec:4-2} except that we set the perturbation size as $\epsilon=\sqrt{0.001\cdot D}$ following \citet{dong2022query} where $D$ is the input dimension.
We compare P-BO ($\lambda^*$) and P-BO ($\lambda=1$) with six strong baselines: NES~\cite{ilyas2018black}, $\mathcal{N}$ATTACK~\cite{li2019nattack}, RGF~\cite{cheng2019improving}, P-RGF~\cite{dong2022query}, Square attack~\cite{andriushchenko2020square}, and BO \cite{ru2019bayesopt}.
We report the attack success rate and the average/median number of queries in Table \ref{tab:l2}.
It can be seen that our P-BO generally leads to higher attack success rates and requires much fewer queries under the $\ell_2$ norm.
This implies that our P-BO method is universal for both $\ell_\infty$ and $\ell_2$ norms.

\begin{table*}[!t]
  \caption{The experimental results of black-box attacks against DenseNet-121, ResNet-50, and SENet-18 under the $\ell_2$ norm on CIFAR-10. We report the attack success rate (ASR) and the average/median number of queries (AVG. Q/MED. Q) needed to generate an adversarial example over successful attacks. We mark the best results in \textbf{bold}.}
  \setlength{\tabcolsep}{4pt}
  \label{tab:l2}
  \centering\small
  \begin{tabular}{l|ccc|ccc|ccc}
    \hline
    \multirow{2}{*}{Method} & \multicolumn{3}{c|}{ResNet-50} & \multicolumn{3}{c|}{DenseNet-121} & \multicolumn{3}{c}{SENet-18}\\
    \cline{2-10}
    & ASR & AVG. Q & MED. Q & ASR & AVG. Q & MED. Q & ASR & AVG. Q & MED. Q \\
    \hline
    NES~\cite{ilyas2018black} & 96.5\% & 335 & 306 & 96.8\% & 321 & 255 & 96.9\% & 307 & 255 \\
    Bandits\textsubscript{T}~\cite{ilyas2018prior} & 97.3\% & 190 & 116 & 98.1\% & 165 & 96 & 98.3\% & 142 & 86 \\
    $\mathcal{N}$ATTACK~\cite{li2019nattack} & 99.2\% & 249 & 204 & 98.4\% & 246 & 204 & 99.9\% & 221 & 203 \\
    Square~\cite{andriushchenko2020square} & 98.6\% & 117 & 56 & 98.1\% & 123 & 62 & 98.7\% & 103 & 49 \\
    RGF~\cite{cheng2019improving} & 99.7\% & 137 & 89 & 98.9\% & 122 & 65 & 99.3\% & 115 & 71 \\
    P-RGF~\cite{dong2022query} & 99.9\% & 30 & 21 & 99.9\% & 28 & 21 & 99.9\% & 27 & 21 \\
    BO~\cite{ru2019bayesopt} & 99.9\% & 98 & 62 & 99.9\% & 80 & 46 & 99.5\% & 72 & 42 \\
    P-BO ($\lambda=1$, \textbf{ours}) & \bf100.0\% & 13 & \bf11 & \bf100.0\% & \bf12 & \bf11 & \bf100.0\% & 13 & \bf11 \\
    P-BO ($\lambda^*$, \textbf{ours}) & \bf100.0\% & \bf12 & \bf11 & \bf100.0\% & \bf12 & \bf11 & \bf100.0\% & \bf12 & \bf11 \\
    \hline
  \end{tabular}
\end{table*}

\subsection{Experimental Results on Different Surrogate Models}
We further conduct experiments under the $\ell_\infty$ norm on CIFAR-10 and ImageNet using different surrogate models.
For CIFAR-10, We adopt VGG-16~\cite{simonyan2014very}, EfficientNet-B0~\cite{tan2019efficientnet}, and TRADES~\cite{zhang2019theoretically} as the surrogate models.
The other experimental settings are the same as those in Sec.~\ref{sec:4-2}.
We report the results of P-BO ($\lambda^*$) and P-BO ($\lambda=1$) in Table~\ref{tab:diff_models_cifar}.
It is worth noting that adopting TRADES as the alternative surrogate model is not a preferable choice. In this scenario, although our P-BO ($\lambda^*$) can adaptively optimize $\lambda^*$ to achieve performance surpassing the BO baseline, it still significantly lags behind the efficiency of standard trained surrogate models. This is primarily due to the substantial disparity in loss landscapes between adversarially trained models and standard trained models, resulting in low similarity between them.
For ImageNet, We use ResNeXt-50~\cite{xie2017aggregated}, ResNet-50~\cite{he2016deep}, and Swin Transformer~\cite{liu2021swin} as the surrogate models.
The other experimental settings are the same as those in Sec.~\ref{sec:4-3}.
We report the results P-BO ($\lambda^*$) and P-BO ($\lambda=1$) in Table~\ref{tab:diff_models_imagenet}.
Compared with the results in Table~\ref{tab:cifar-l8} and Table~\ref{tab:imagenet-l8} in the paper, it can be seen that our P-BO generally leads to higher attack success rates and requires much fewer queries using different surrogate models.
This indicates that our P-BO is not sensitive to the selection of surrogate models, enabling consistent improvements in both the attack success rate and query efficiency across different surrogate models.

\begin{table*}[!t]
  \caption{The experimental results of black-box attacks under the $\ell_\infty$ norm on CIFAR-10 using different surrogate models. We report the attack success rate (ASR) and the average/median number of queries (AVG. Q/MED. Q) needed to generate an adversarial example over successful attacks.}
  \setlength{\tabcolsep}{4pt}
  \label{tab:diff_models_cifar}
  \centering\small
  \begin{tabular}{l|l|ccc|ccc|ccc}
    \hline
    \multirow{2}{*}{Surrogate Model} & \multirow{2}{*}{Method} & \multicolumn{3}{c|}{ResNet-50} & \multicolumn{3}{c|}{DenseNet-121} & \multicolumn{3}{c}{SENet-18}\\
    \cline{3-11}
    & & ASR & AVG. Q & MED. Q & ASR & AVG. Q & MED. Q & ASR & AVG. Q & MED. Q \\
    \hline
    - & BO~\cite{ru2019bayesopt} & 99.6\% & 83 & 44 & 99.7\% & 93 & 51 & 99.7\% & 81 & 41 \\
    \hline
    \multirow{2}{*}{WRN-34-10} & P-BO ($\lambda=1$) & 99.9\% & 16 & 11 & 99.7\% & 25 & 11 & 99.9\% & 14 & 11 \\
    & P-BO ($\lambda^*$) & 100.0\% & 15 & 11 & 100.0\% & 19 & 11 & 100.0\% & 14 & 11 \\
    \hline
    \multirow{2}{*}{VGG-16} & P-BO ($\lambda=1$) & 99.4\% & 25 & 11 & 99.9\% & 23 & 11 & 99.6\% & 19 & 11 \\
    & P-BO ($\lambda^*$) & 100.0\% & 20 & 11 & 100.0\% & 17 & 11 & 100.0\% & 16 & 11 \\
    \hline
    \multirow{2}{*}{EfficientNet-B0} & P-BO ($\lambda=1$) & 99.5\% & 21 & 11 & 99.6\% & 19 & 11 & 99.7\% & 17 & 11 \\
    & P-BO ($\lambda^*$) & 100.0\% & 19 & 11 & 100.0\% & 16 & 11 & 100.0\% & 17 & 11 \\
    \hline
    \multirow{2}{*}{TRADES} & P-BO ($\lambda=1$) & 99.7\% & 84 & 36 & 99.7\% & 73 & 33 & 99.0\% & 66 & 31 \\
    & P-BO ($\lambda^*$) & 99.9\% & 66 & 34 & 99.8\% & 59 & 32 & 99.9\% & 62 & 32 \\
    \hline
  \end{tabular}
\end{table*}

\begin{table*}
  \caption{The experimental results of black-box attacks under the $\ell_\infty$ norm on ImageNet using different surrogate models. We report the attack success rate (ASR) and the average/median number of queries (AVG. Q/MED. Q) needed to generate an adversarial example over successful attacks. The subscript ``D'' denotes the methods with dimensionality reduction.}
  \label{tab:diff_models_imagenet}
\setlength{\tabcolsep}{4.25pt}
  \centering\small
  \begin{tabular}{l|l|ccc|ccc|ccc}
    \hline
    \multirow{2}{*}{Surrogate Model} & \multirow{2}{*}{Method} & \multicolumn{3}{c|}{Inception-v3} & \multicolumn{3}{c|}{MobileNet-v2} & \multicolumn{3}{c}{ViT-B/16}\\
    \cline{3-11}
    & & ASR & AVG. Q & MED. Q & ASR & AVG. Q & MED. Q & ASR & AVG. Q & MED. Q \\
    \hline
    - & BO\textsubscript{D}~\cite{ru2019bayesopt} & 94.1\% & 104 & 58 & 98.2\% & 102 & 61 & 86.7\% & 170 & 116 \\
    \hline
    \multirow{2}{*}{ResNet-152} & P-BO\textsubscript{D} ($\lambda=1$) & 85.4\% & 182 & 90 & 86.8\% & 193 & 56 & 67.5\% & 236 & 240 \\
    & P-BO\textsubscript{D} ($\lambda^*$) & 94.4\% & 81 & 45 & 98.8\% & 94 & 60 & 88.2\% & 148 & 81 \\
    \hline
    \multirow{2}{*}{ResNeXt-50} & P-BO\textsubscript{D} ($\lambda=1$) & 81.8\% & 265 & 248 & 87.5\% & 231 & 57 & 64.4\% & 264 & 235 \\
    & P-BO\textsubscript{D} ($\lambda^*$) & 96.4\% & 92 & 50 & 98.3\% & 84 & 51 & 89.5\% & 157 & 91 \\
    \hline
    \multirow{2}{*}{ResNet-50} & P-BO\textsubscript{D} ($\lambda=1$) & 72.9\% & 264 & 270 & 79.4\% & 206 & 31 & 58.7\% & 304 & 294 \\
    & P-BO\textsubscript{D} ($\lambda^*$) & 95.3\% & 91 & 51 & 98.5\% & 91 & 60 & 89.3\% & 165 & 87 \\
    \hline
    \multirow{2}{*}{Swin Transformer} & P-BO\textsubscript{D} ($\lambda=1$) & 84.5\% & 309 & 242 & 89.8\% & 160 & 67 & 76.4\% & 269 & 137 \\
    & P-BO\textsubscript{D} ($\lambda^*$) & 97.6\% & 90 & 59 & 99.4\% & 90 & 49 & 90.4\% & 152 & 64 \\
    \hline
  \end{tabular}
\end{table*}

\subsection{Experimental Results Compared with $\pi$-BO}
$\pi$-BO~\cite{hvarfner2022pi} incorporates prior beliefs about the location of the optimum in the form of a probability distribution on the acquisition function. We extend our experiments under the $\ell_\infty$ norm on CIFAR-10 using $\pi$-BO. We employ a Gaussian distribution as the probability distribution, with the mean derived from the optimal point obtained through PGD attack on the surrogate model and a constant variance that encapsulates prior beliefs regarding the optimal location. The remaining experimental configurations remain consistent with those outlined in Section~\ref{sec:4-2}. We compare $\pi$-BO with BO and P-BO in Table~\ref{tab:pibo}.
While $\pi$-BO exhibits some improvements over BO, there remains a significant performance gap compared to our P-BO method. This is because the surrogate model is more appropriately utilized as a function prior and P-BO can effectively integrate the prior information for black-box attacks.

\begin{table*}[!t]
  \caption{The experimental results of black-box attacks compared with $\pi$-BO under the $\ell_\infty$ norm on CIFAR-10. We report the attack success rate (ASR) and the average/median number of queries (AVG. Q/MED. Q) needed to generate an adversarial example over successful attacks.}
  \setlength{\tabcolsep}{4pt}
  \label{tab:pibo}
  \centering\small
  \begin{tabular}{l|ccc|ccc|ccc}
    \hline
    \multirow{2}{*}{Method} & \multicolumn{3}{c|}{ResNet-50} & \multicolumn{3}{c|}{DenseNet-121} & \multicolumn{3}{c}{SENet-18}\\
    \cline{2-10}
    & ASR & AVG. Q & MED. Q & ASR & AVG. Q & MED. Q & ASR & AVG. Q & MED. Q \\
    \hline
    BO~\cite{ru2019bayesopt} & 99.6\% & 83 & 44 & 99.7\% & 93 & 51 & 99.7\% & 81 & 41 \\
    $\pi$-BO~\cite{hvarfner2022pi} & 99.9\% & 78 & 12 & 99.9\% & 61 & 11 & 99.4\% & 49 & 11 \\
    P-BO ($\lambda^*$, \textbf{ours}) & 100.0\% & 15 & 11 & 100.0\% & 19 & 11 & 100.0\% & 14 & 11 \\
    \hline
  \end{tabular}
\end{table*}

\subsection{Ablation Study}

Here, we conduct an ablation study to validate the necessity of Bayesian optimization (BO). We directly employ the function prior as the acquisition function $\alpha$ in Algorithm \ref{alg:pg-bayesopt}, independent of observed values from the target function, denoted as PBO w/o BO. This is equivalent to repeatedly conducting transfer attacks until succeeded. We adopt the experimental setup identical to Sec.~\ref{sec:4-3} on ImageNet, and the results for the PBO w/o BO are presented in Table \ref{tab:imagenet-l8-2}. It can be observed that this approach also demonstrates promising performance, often achieving successful attacks with only a few queries. However, the success rate is significantly lower compared to the P-BO algorithm. This suggests that the integration of function prior and observed values from the target function for optimization exploration, as employed by Bayesian optimization, is a crucial and effective approach.

\begin{table*}[!t]
  \caption{The experimental results of P-BO variants against Inception-v3, MobileNet-v2, and ViT-B/16 under the $\ell_\infty$ norm on ImageNet. We report the attack success rate (ASR) and the average/median number of queries (AVG. Q/MED. Q) needed to generate an adversarial example over successful attacks. We mark the best results in \textbf{bold}. The subscript ``D'' denotes the methods with dimensionality reduction.}
  \label{tab:imagenet-l8-2}
  \centering\small
  \begin{tabular}{l|ccc|ccc|ccc}
    \hline
    \multirow{2}{*}{Method} & \multicolumn{3}{c|}{Inception-v3} & \multicolumn{3}{c|}{MobileNet-v2} & \multicolumn{3}{c}{ViT-B/16}\\
    \cline{2-10}
    & ASR & AVG. Q & MED. Q & ASR & AVG. Q & MED. Q & ASR & AVG. Q & MED. Q \\
    \hline
    P-BO\textsubscript{D} w/o BO & 39.8\% & 55 & 2 & 64.8\% & 102 & 11 & 25.6\% & 96 & 6 \\
    P-BO\textsubscript{D} ($\lambda=1$) & 85.4\% & 182 & 90 & 86.8\% & 193 & 56 & 67.5\% & 236 & 240 \\
    P-BO\textsubscript{D} ($\lambda^*$) & 94.4\% & 81 & 45 & 98.8\% & 94 & 60 & 88.2\% & 148 & 81 \\
    \hline
  \end{tabular}
\end{table*}

\end{document}